\documentclass{article}

\usepackage{arxiv}

\usepackage[utf8]{inputenc} 
\usepackage[T1]{fontenc}    
\usepackage{hyperref}       
\usepackage{url}            
\usepackage{booktabs}       
\usepackage{amsfonts}       
\usepackage{nicefrac}       
\usepackage{microtype}      
\usepackage{lipsum}

\usepackage[round]{natbib}
\usepackage{graphicx}
\usepackage{pbox} 
\usepackage{wrapfig}
\usepackage{algorithm}
\usepackage{algpseudocode}
\usepackage{amsmath}
\usepackage{bm}   
\usepackage{amsfonts}
\usepackage{adjustbox}
\usepackage{float}
\usepackage[caption = false]{subfig}
\usepackage{xcolor}

\title{Improving Coordination in Small-Scale Multi-Agent Deep Reinforcement Learning through Memory-driven Communication}
\author{
    Emanuele Pesce \\
	WMG,University of Warwick\\
	Coventry, CV4 7AL\\
	\texttt{e.pesce@warwick.ac.uk} \\
  \And
    Giovanni Montana \\
	WMG,University of Warwick\\
	Coventry, CV4 7AL\\
	\texttt{g.montana@warwick.ac.uk} \\
}

\begin{document}
	\maketitle	
	\begin{abstract}
		Deep reinforcement learning algorithms have recently been used to train multiple interacting agents in a centralised manner whilst keeping their execution decentralised. When the agents can only acquire partial observations and are faced with task{s} requiring coordination and synchronisation skills, inter-agent communication plays an essential role. In this work, we propose a framework for multi-agent training using deep deterministic policy gradients that enables concurrent, end-to-end learning of an explicit communication protocol through a memory device. During training, the agents learn to perform read and write operations enabling them to infer a shared representation of the world. We empirically demonstrate that concurrent learning of the communication device and individual policies can improve inter-agent coordination and performance in small-scale systems. {\color{black}Our experimental results show that the proposed method achieves superior performance in scenarios with up to six agents.}
		We illustrate how different communication patterns can emerge {on six} different tasks of {increasing complexity}. {Furthermore, we study the effects of corrupting the communication channel, provide a visualisation of the time-varying memory content as the underlying task is being solved and validate the building blocks of the proposed memory device through ablation studies.}
	\end{abstract}
	
	\section{Introduction}
	
	Reinforcement Learning (RL) allows agents to learn how to map observations to actions through feedback reward signals \citep{sutton1998introduction}. Recently, deep neural networks \citep{lecun2015deep,schmidhuber2015deep} have had a noticeable impact on RL \citep{li2017deep}. {\color{black}T}hey provide flexible models for learning value functions and policies{\color{black},} allow to overcome difficulties related to large state spaces, and eliminate the need for hand-crafted features and ad-hoc heuristics \citep{cortes2002coverage,parker2003multi,olfati2007consensus}. Deep reinforcement learning ({\color{black}DRL}) algorithms{\color{black}, which usually {\color{black}rely} on deep neural networks to approximate functions,} have been successfully {\color{black}employed} in single-agent systems, including video game playing \citep{mnih2015human}, robot locomotion \citep{lillicrapHPHETS15}, object localisation \citep{caicedo2015active} and data-center cooling \citep{DeepmindCooling}. 
	
	Following the uptake of DRL in single-agent domains, there is now a need to develop {\color{black}improved} learning algorithms for multi-agent (MA) systems {\color{black}where} additional challenges {\color{black}arise}. Markov Decision Processes, upon which DRL methods rely, assume that the reward distribution and dynamics are stationary \citep{hernandez2017survey}. When multiple learners interact with each other, this property is violated because the reward that an agent receives also depends on other agents' actions \citep{laurent2011world}. This issue, known as the {\it moving-target} problem \citep{tuyls2012multiagent}, removes convergence guarantees and introduces additional learning instabilities. Further difficulties arise from environments characterized by partial observability \citep{singh1994learning,chu2017parameter,peshkin2000learning} whereby the agents do not have full access to the world state, and where coordination skills are essential.
	
	An important challenge in multi-agent DRL is how to facilitate communication {\color{black}amongst} interacting agents. Communication is widely known to play a critical role in promoting coordination between humans \citep{szamado2010pre}.  
	{\color{black} Humans have been proven to excel at communicating even in absence of a conventional code \citep{de2010exploring}.}
	When coordination is required and no common languages exist, simple communication protocols are likely to emerge \citep{selten2007emergence}.
	{\color{black} Human communication involves more than sending and receiving messages, it requires specialized interactive intelligence where receivers have the ability to recognize intentions and senders can properly design messages \citep{wharton2003natural}. 
		The emergence of communication has been widely investigated \citep{garrod2010can,theisen2010systematicity}, for example new signs and symbols can emerge when it comes to represent real concepts. \citet{fusaroli2012coming} demonstrated that language can be seen as a social coordination device learnt through reciprocal interaction with the environment for optimizing coordinative dynamics.}
	The relation between communication and coordination has been widely discussed \citep{vorobeychik2017does, demichelis2008language,miller2004communication,kearns2012experiments}.
	{\color{black} Communication is an essential skill in many tasks: for instance, in critical situations, where is of fundamental importance to properly manage critical and urgent situations, like emergency response organizations \citep{comfort2007crisis}.
		In multiplayer videogames, it is often essential to reach a sufficiently high level of coordination required to succeed \citep{chen2009communication}.} 
	
	{\color{black}Two-agents systems have often been studied when looking at the effects of communication on coordination. \citet{galantucci2005experimental} showed that humans can easily produce new protocols to overcame the lack of a common language, through experiments in which pairs of participants playing video games were allowed to send messages through a medium that prevented the use of standard symbols.}
	In two-players game{s\color{black}, like the Battle of the Sexes}, improved coordination resulted when allowing gamers to exchange messages 	\citep{ccooper1989communiation}. 
	{\color{black}Human conversations can be interpreted as a bi-directional communication form, where the same entity can both send and receive messages \citep{lasswell1948structure}. This kind of communication can be efficiently explored in small-scale systems through coordination games \citep{cooper1992forward} and often it is the key to achieve success in real world scenarios such as bargaining with incomplete information \citep{brosig2003information}.}
	
	
	{\color{black}
		Analogously, the importance of communication has been recognised when designing artificial MA learning systems, especially in tasks requiring synchronization \citep{scardovi2008synchronization,wen2012consensus}. For example, in navigation tasks, agents can localise each other more easily through shared information \citep{fox2000probabilistic}. In group strategy coordination, such as automating negotiations, communication is fundamental to improve the final outcome \citep{wunder2009communication,ito2011innovations}.  }
	A wide range of MA applications have benefitted from inter-agent message passing including distributed smart grid control \citep{pipattanasomporn2009multi}, consensus in networks \citep{you2011network}, multi-robot control \citep{ren2008distributed}, autonomous vehicle driving \citep{petrillo2018adaptive}, elevators control \citep{crites1998elevator}, soccer-playing robots \citep{stone1998towards} {\color{black} and for language learning in two-agent systems \citep{lazaridou2016multi}}.

	Recently, \citet{lowe2017multi} have proposed MADDPG (Multi-Agent Deep Deterministic Policy Gradient). Their approach extends the actor-critic algorithm \citep{degris2012off} in which each agent has an actor to select actions and a critic to evaluate them. MADDPG embraces the centralised learning and decentralised execution paradigm (CLDE) \citep{foerster2016learning, kraemer2016multi, oliehoek2007q}. During centralised training, the critics receive observations and actions from all the agents whilst the actors only see their local observations. On the other hand, the execution only relies on actors. This approach has been designed to address the emergence of environment non-stationarity \citep{tuyls2012multiagent, laurent2011world} and has been shown to perform well in a number of mixed competitive and cooperative environments. In MADDPG, the agents can only share each other's actions and observations during training through their critics, but do not have the means to develop an explicit form of communication through their experiences. {\color{black} The input size of each critic increases linearly with the number of agents \citep{lowe2017multi}, which hinders its scalability \citep{jiang2018learning}.} 
	
	In this article, we consider tasks requiring strong coordination and synchronization skills. { In order to thoroughly study the effects of communication on these scenarios, we focus on small-scale systems. 
	This allows us to design tasks with an increasing level of complexity, and simplifies the investigation of possible correlations between the level of messages being exchanged and any environmental changes. 
	{\color{black} We provide empirical evidences that the proposed method reaches very good performance on a range of two-agent scenarios when a high level of cooperation is required. In addition, we present experimental results for systems with up to six agents in the Supplementary Material (Section \ref{subsec:moreagents_CN} and \ref{subsec:moreagents_POCN}).}
	In the real world, there is range of applications involving the coordination of only a few actors, for example motor interactions like sawing or cooperative lifting of heavy loads \citep{jarrasse2012framework}}.

	In such cases, being able to communicate information beyond the private observations, and infer a shared representation of the world through interactions, becomes essential. Ideally, an agent should be able to remember its current and past experience generated when interacting with the environment, learn how to compactly represent these experiences in an appropriate manner, and share this information for others to benefit from. Analogously, an agent should be able to learn how to decode the information generated by other agents and leverage it under every environmental state. 
	The above requirements are captured here by introducing a communication mechanism facilitating information sharing within the CLDE paradigm. Specifically, we provide the agents with a shared communication device that can be used to learn from their collective private observations and share relevant messages with others. Each agent also learns how to decode the memory content in order to improve its own policy. Both the read and write operations are implemented as parametrised, non-linear gating mechanisms that are learned concurrently with the individual policies. When the underlying task to be solved demands for complex coordination skills, we demonstrate that our approach can achieve higher performance compared to the MADDPG baseline {\color{black} in small-scale systems}. Furthermore, we demonstrate that being able to learn end-to-end a communication protocol jointly with the policies can also improve upon a {\it meta-agent} approach whereby all the agents perfectly share all their observations and actions in both training and execution. We {\color{black} investigate a potential interpretation} of the communication patterns that have emerged when training two-agent systems through time-varying low-dimensional projections and their visual assessment, and demonstrate how these patterns correlate with the underlying tasks being learned.

	{\color{black}This article is organised as follow. In Section \ref{sec:relatedwork} a general overview of related work is offered to characterize state-of-the-art approaches for MARL with special focus on communication systems. Section \ref{sec:mdmaddpg} contains the formalization of the problem setup, the details of the proposed method and the description of the learning process; all the experiments are reported in Section \ref{sec:experiments} where results are presented in terms of numerical metrics to evaluate the performance achieved on six different scenarios; an analysis of the communication channel is provided to support qualitative insights of the exchanged messages. Conclusive comments are given in Section \ref{sec:conclusions}. In the Supplementary Material, Section \ref{sec:metaagent} describes details of MA-MADDPG, a comparative method, and Section \ref{sec:additional_exp} presents a range of additional experiments to further investigate the effects of memory corruption; changes in performance when increasing the number of agents; an ablation study to validate the components used in the proposed method; box plots with the main results, an assessment of the robustness of the method when changing the random seeds; additional analyses of the communication channel.}
	
	\section{Related Work}\label{sec:relatedwork}
	
	
	The problem of reinforcement learning in cooperative environments has been  studied extensively \citep{littman1994markov,schmidhuber1996general,panait2005cooperative,matignon2007hysteretic}. Early attempts exploited single-agent techniques like Q-learning to train all agents independently \citep{tan1993multi}, but suffered from the excessive size of the state space resulting from having multiple agents. Subsequent improvements were obtained using variations of Q-learning \citep{ono1996multi, guestrin2002coordinated} and distributed approaches \citep{lauer2000algorithm}. 
	More recently, DRL techniques like DQN \citep{mnih2013playing} have led to superior performance in single-agents settings by approximating policies through deep neural networks. 
	\cite{tampuu2017multiagent} have demonstrated that an extension of the DQN is able to train multiple agents independently to solve {\color{black} a popular two-agent system}, the Pong game.
	\cite{gupta2017cooperative} have analyzed the performance of popular DRL algorithms, including DQN, DDPG \citep{lillicrapHPHETS15}, TRPO \citep{schulman2015trust} and actor-critic on different MA environments, and have introduced a curriculum learning approach to increase scalability. \cite{foerster2017counterfactual} have suggested using a centralized critic for all agents that marginalises out a single's agent action while other agents' actions are kept fixed.  	\citet{iqbal2018actor} proposed MAAC (Multi-Actor-Attention-Critic), a framework {\color{black}for learning} decentralised policies with centralised critics, which selects relevant information for each agent at every time-step through an attention mechanism. {\color{black} In more recent work, a probabilistic recursive reasoning framework has been proposed to model behaviours in a two-agents context; each agent, through variational Bayes methods, approximates the other agent policy to predict its strategy and then to improve its own policy \citep{wen2019probabilistic}.}
	

	The role of communication in cooperative settings has also been explored, and different methods have been proposed differing on how the communication channels have been formulated using DRL techniques. Many approaches rely on {\it implicit} communication mechanisms whereby the weights of the neural networks used to implement policies or action-value functions are shared across agents or modelled to allow inter-agent information flow. For instance, in CommNet \citep{sukhbaatar2016learning}, the policies are implemented through subsets of units of a large feed-forward neural network mapping the inputs of all agents to actions. At any given time step, the hidden states of each agent are used as messages, averaged and sent as input for the next layer. \citet{singh2018learning} proposed IC3NEt, a model designed to improve CommNet, where the hidden states of the agents are also used as messages, but this time they are averaged only after being weighted by a gating mechanism. In addition, in IC3Net, each agent is implemented through an { Long Short Term Memory (LSTM)} \citep{hochreiter1997long} {\color{black}in order to} consider the history of the seen observations.
	In BiCNet \citep{peng2017multiagent}, the agents' policies and value networks are connected through bidirectional neural networks, and trained using an actor-critic approach. 
	\cite{jiang2018learning} proposed an attention mechanism that, when a need for communication emerges, selects which subsets of agents should communicate; the hidden states of their policy networks are integrated through an LSTM to generate a message that is used as input for the next layer of the policy network. \citet{das2018tarmac} utilised a soft attention mechanism to allow the agents to select the recipients of their messages. Each agent, along with the message, broadcasts a signature which can be used to encode agent-specific information.
	\cite{kong2017revisiting} introduce a master-slave architecture whereby a master agent provides high-level instructions to organize the slave agents in an attempt to achieve fine-grained optimality. 
	{\color{black} Similarly, in Feudal Multiagent Hierarchies \citep{ahilan2019feudal}, an agent acts as \textquotedblleft manager\textquotedblright and learns to communicate sub-goals to multiple workers operating simultaneously. A different approach is instead provided by the Bayesian Action Decoder (BAD)  \citep{foerster2018bayesian}, a technique for two-agent settings where an approximate Bayesian update is used to produce public belief that directly conditions the actions of all agents.}
	
	In our work, we introduce a mechanism to generate {\it explicit} messages capturing relevant aspects of the world, which the agents are able to collectively learn using their observations and interactions. The messages are then sent and received to complement their private observations when making decisions. Some aspects of our approach are somewhat related to DIAL (Differentiable Inter-Agent Learning) \citep{foerster2016learning} in that the communication is enabled by differentiable channels allowing the gradient of the Q-function to bring the proper feedback {\color{black} in small-scale scenarios}.  Like DIAL, we would like the agents to share explicit messages. However, whereas DIAL uses simple and pre-determined protocols, {\color{black}our} agents {\color{black}are given} the ability to infer complex protocols from experience, without necessarily relying on pre-defined ones, and utilise those to learn better policies. 
	Explicit messages are also used in SchedNet \citep{kim2018learning} to investigate situations where the bandwidth is limited and only some of the agents are allowed to communicate. In their approach, {\color{black}also focusing on small-case scenarios to better capture the scheduling constraints,} the agents produce messages by encoding their observations and a scheduler decides whether an agent is allowed to use a communication channel. A limited bandwidth channel is also used in our work, but all the agents have full access to the channel.

	\section{Memory-driven MADDPG} \label{sec:mdmaddpg}
	
	\subsection{Problem setup}
	We consider a system with $N$ interacting agents, {\color{black} where $N$ is typically small}, and adopt a multi-agent extension of partially observable Markov decision processes \citep{littman1994markov}. This formulation assumes a set, $ \mathcal{S}$, containing all the states characterising the environment; a sequence $\{\mathcal{A}_1, \mathcal{A}_2, \dots, \mathcal{A}_N\}$ where each $\mathcal{A}_i$ is a set of possible actions for the $i^{th}$ agent; a sequence $\{\mathcal{O}_1, \mathcal{O}_2, \dots, \mathcal{O}_N\}$ where each $ \mathcal{O}_i$ contains the observations available to the $i^{th}$ agent. Each $\bm{o}_i \in \mathcal{O}_i$ provides a partial characterisation of the current state and is private for that agent. Every action $a_i \in \mathcal{A}_i$ is deterministically chosen accordingly to a policy function, $ \bm{\mu}_{\theta_i}:  \mathcal{O}_i \mapsto A_i $, parametrised by $\theta_i$. The environment generates a next state according to a transition function, $ \mathcal{T}: S \times \mathcal{A}_1 \times \mathcal{A}_2 \times \dots \times \mathcal{A}_N  $, that considers the current state and the $N$ actions taken by the agents.
	The reward received by an agent, $ r_i : \mathcal{S} \times \mathcal{A}_1 \times \mathcal{A}_2 \times \dots \times \mathcal{A}_N  \mapsto \mathbb{R}$ is a function of states and actions. Each agent learns a policy that maximises the expected discounted future rewards over a period of $T$ time steps, $J(\theta_i)  = \mathbb{E} [R_i]$, where $R_i  = \sum_{t=0}^{T} \gamma^t r_i(s^t_i,a^t_i)$ is the  $\gamma$-discounted sum of future rewards. During training, we would like an agent to learn by using not only its own observations, but through a collectively learned representation of the world that accumulates through experiences coming from all the agents. At the same time, each agent should develop the ability to interpret this shared knowledge in its own unique way as needed to optimise its policy. Finally, the information sharing mechanism would need to be designed in such a way to be used in both training and execution.

	\subsection{Memory-driven communication} \label{sec:sharedmemory}
	
\begin{figure}
		\includegraphics[scale=0.5]{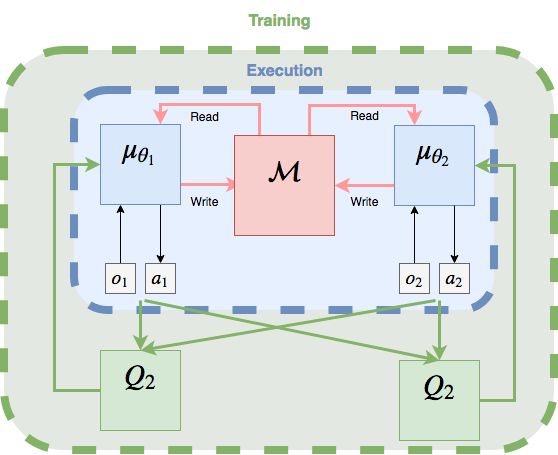}
		\caption{The MD-MADDPG framework. { During training and testing, each policy uses its observation and the content of the shared memory to produce a new action and then update the shared channel. Critics are used during training only and each one of them takes {\color{red} as} input all the observations and actions.}}\label{fig:architecture} 
	\end{figure}

	We introduce a shared communication mechanism enabling agents to establish a communication protocol through a memory device $ \mathcal{M} $ of {\color{black}pre-determined} capacity $M$ (Figure \ref{fig:architecture}). The device is designed to store a message $  \mathbf{m} \in \mathbb{R}^M $ which progressively captures the collective knowledge of the agents as they interact. An agent's policy becomes $ \bm{\mu}_{\theta_i}: \mathcal{O}_i \times \mathcal{M} \mapsto A_i $, i.e. it is dependent on the agent's private observation as well as the collective memory. Before taking an action, each agent accesses the memory device to initially retrieve and interpret the message left by others. After reading the message, the agent performs a writing operation that updates the memory content. During training, these operations are learned without any \emph{a priori} constraint on the nature of the messages other than the device's size, $M$. During execution, the agents use the communication protocol that they have learned to read and write the memory over an entire episode. We aim to build a model trainable \emph{end-to-end} only through reward signals, and use neural networks as function approximators for policies, and learnable gated functions {\color{black} as mechanisms to facilitate} an agent's interactions with the memory. The chosen parametrisations of these operations are presented and discussed below. 
	
	\paragraph{Encoding operation.}
	
	Upon receiving its private observations, each agent maps them on to an embedding representing the agent's current vision of the state:
	\begin{equation}\label{eq:encoding}
	\mathbf{e}_i = \varphi_{\theta_{i}^e}^{enc}(\bm{o}_i), \hspace{1cm} \mathbf{e}_i \in \mathbb{R}^{E}
	\end{equation}
	where $ \varphi_{\theta_{i}^e}^{enc} $ is a neural network parametrised by $ \theta_{i}^e $. The embedding $ \mathbf{e}_i $ plays a fundamental role in selecting a new action and in the reading and writing phases. 
	
	\paragraph{Read operation.}
	
	After encoding the current information, the agent performs a read operation allowing to extract and interpret relevant knowledge that has been previously captured through $\mathcal{M}$. By interpreting this information content, the agent has access to what other agents have learned. A context vector $\mathbf{h}_i$ is generated to capture spatio-temporal information previously encoded in $\mathbf{e}_i$ through a linear mapping, 
	\begin{equation*} \label{eq:readh}
	\mathbf{h}_i = \mathbf{W}_i^{h}\mathbf{e}_i, \hspace{1cm} \mathbf{h}_i \in \mathbb{R}^{H}, \mathbf{W}_i^{h} \in \mathbb{R}^{H \times E}
	\end{equation*}
	where $ \mathbf{W}_i^{h} $ represent the learnable weights of the linear projection. While $\mathbf{e}_i$ is defined as general observation encoder, $\mathbf{h}_i$ is specifically designed to extract features for the reading operation. {\color{black} The context vector $\mathbf{h}_i$ can be interpreted as an agent's internal representation that uses the observation embedding $\mathbf{e}_i$ to extract information to be utilized by the gating mechanism only (Eq. \ref{eq:readk}); its output is then used to extract information from the memory. 
		The main function of the context vector is to facilitate the emergence of an internal representation specifically designed for interpreting the memory content during the read phase.
		An ablation study aimed at investigating the added benefits introduced by $\mathbf{h}_i$ is provided in the Supplementary Material (\ref{subsec:ablatation_context}). This study supports our intuition that the context vector is crucial for the proper functioning of the entire framework on more complex environments.} The agent observation embedding $\mathbf{e}_i$, the reading context vector $\mathbf{h}_i$ and the current memory $\mathbf{m}$ contain different types of information that are used jointly as inputs to learn a gating mechanism,
	\begin{equation}\label{eq:readk}
	\mathbf{k}_i = \sigma (\mathbf{W}_i^{k}[\mathbf{e}_i,\mathbf{h}_i, \mathbf{m}]),  \hspace{1cm} \mathbf{k}_i \in [0,1]^{M},\mathbf{W}_i^{k} \in \mathbb{R}^{M \times (E + H + M)}
	\end{equation}
	where $ \sigma( \cdot ) $ is the sigmoid function and $[\mathbf{e}_i,\mathbf{h}_i, \mathbf{m}]$ means that the three vectors are concatenated. The values of $ \mathbf{k}_i $ are used as weights to modulate the memory content and extract the information from it, i.e. 
	\begin{equation}\label{eq:readv}
	\mathbf{r}_i =\mathbf{m} \odot \mathbf{k}_i
	\end{equation} 
	where $ \odot $ represents the Hadamard product. $\mathbf{k}_i$ takes values in $[0,1]$ and its role is to potentially downgrade the information stored in memory or even completely discard the current content. Learning agent-specific weights $\mathbf{W}_i^h$ and $\mathbf{W}_i^k $ means that each agent is able to interpret $\mathbf{m}$ in its own unique way. As the reading operation strongly depends on the current observation, the interpretation of $\mathbf{m}$ can change from time to time depending on what an agent sees during an episode. 
	Given that $\mathbf{r}_i$ depends on $\mathbf{m}$ and $\mathbf{e}_i$ (from $\bm{o}_i$ in Eq. \ref{eq:encoding}), we lump all the adjustable parameters into $\theta_{i}^{\zeta} = \{ \mathbf{W}_i^h, \mathbf{W}_i^k \} $ and write 
	\begin{equation}\label{eq:readsummary}
	\mathbf{r}_i = \zeta_{\theta_{i}^{\zeta}}(\bm{o}_i, \mathbf{m}).
	\end{equation}
	
	\paragraph{Write operation.}
	
	In the writing phase, an agent decides what information to share and how to properly update the content of the memory whilst taking into account the other agents. The write operation is loosely inspired by the LSTM \citep{hochreiter1997long} where the content of the memory is updated through gated functions regulating what information is kept and what is discarded. Initially, the agent generates a candidate memory content, $\mathbf{c}_i$, which depends on its own encoded observations and current shared memory through a non-linear mapping,
	\begin{equation*}
	\mathbf{c}_i =  tanh(\mathbf{W}_i^{c}[\mathbf{e}_i,\mathbf{m}]) \hspace{1cm} \mathbf{c}_i \in [-1,1]^{M},\mathbf{W}_i^{c} \in \mathbb{R}^{M \times (E + M)}
	\end{equation*}
	where $\mathbf{W}_i^{c}$ are weights to learn.
	An input gate, $\mathbf{g}_i$, contains the values used to regulate the content of this candidate while a forget gate, $\mathbf{f}_i$, is used to decide what to keep and what to discard from the current $ \mathbf{m} $. These operations are described as follows:
	\begin{align*}
	\begin{split}
	\mathbf{g}_i = {} & \sigma (\mathbf{W}_i^{g}[\mathbf{e}_i,\mathbf{m}]),   \hspace{1cm} \mathbf{g}_i \in [0,1]^{M},\mathbf{W}_i^{g} \in \mathbb{R}^{M \times (E + M)}\\
	\mathbf{f}_i = {} & \sigma (\mathbf{W}_i^{f}[\mathbf{e}_i,\mathbf{m}]), \hspace{1cm} \mathbf{f}_i \in [0,1]^{M},\mathbf{W}_i^{f} \in \mathbb{R}^{M \times (E + M)}.
	\end{split}
	\end{align*}
	The $i^{th}$ agent then finally generates an updated message as a {\color{black}weighted} linear combination of old and new messages{\color{black},} as follows:
	\begin{equation}\label{eq:writev}
	\mathbf{m'} =  \mathbf{g}_i \odot \mathbf{c}_i + \mathbf{f}_i \odot \mathbf{m}.
	\end{equation}
	The update $\mathbf{m}'$ is stored in memory $\mathcal{M}$ and made accessible to other agents. At each time step, agents sequentially read and write the content of the memory using the above procedure. Since $\mathbf{m'}$ depends on $\mathbf{m}$ and  $\bm{e}_i$ (derived from $\bm{o}_i$ in Eq. \ref{eq:encoding}) we collect all the parameters into $\theta_{i}^{\xi} = \{ \mathbf{W}_i^c, \mathbf{W}_i^g, \mathbf{W}_i^f \}$ and write the writing operation as: 
	\begin{equation}\label{eq:writesummary}
	\mathbf{m'} = \xi_{\theta_{i}^{\xi}}(\bm{o}_i, \mathbf{m}).
	\end{equation}
	\paragraph{Action selector.}
	Upon completing both read and write operations, the agent is able to take an action, $a_i$, which depends on the current encoding of its observations, its own interpretation of the current memory content and its updated version, that is
	\begin{equation}\label{eq:actionselector}
	a_i =  \varphi_{\theta_{i}^a}^{act}(\mathbf{e}_i,\mathbf{r}_i,\mathbf{m'})
	\end{equation}
	where $ \varphi_{\theta_{i}^a}^{act} $ is a neural network parametrised by ${\theta_{i}^a}$. The resulting policy function can be written as a composition of functions:
	\begin{equation}\label{eq:policy}
	\bm{\mu}_{\theta_i}(\bm{o}_i, \mathbf{m}) = \varphi_{\theta_{i}^a}^{act}(\varphi_{\theta_{i}^e}^{enc}(\bm{o}_i),\zeta_{\theta_{i}^{\zeta}}(\bm{o}_i, \mathbf{m}), \xi_{\theta_{i}^{\xi}}(\bm{o}_i, \mathbf{m})) 
	\end{equation}
	in which $\theta_{i} = \{ \theta_i^a, \theta_i^e, \theta_i^\zeta, \theta_i^\xi \}$ contains all the relevant parameters.

	\paragraph{Learning algorithm.}
	All the agent-specific policy parameters, i.e.  $\theta_{i}$, are learned \emph{end-to-end}.
	We adopt an actor-critic model within a CLDE framework \citep{foerster2016learning,lowe2017multi}. In the standard actor-critic model \citep{degris2012off}, {\color{black}we have}  an actor to select the actions, and a critic, to evaluate the actor moves {\color{black}and} provide feedback. In DDPG \citep{silver2014deterministic,lillicrapHPHETS15}, neural networks are used to approximate both the actor, represented by the policy function
	$\bm{\mu}_{\omega_i}$, and its corresponding critic, represented by an action-value function $ Q^{\bm{\mu}_{\omega_i}}: \mathcal{O}_i \times \mathcal{A}_i \mapsto \mathbb{R} $, in order to maximize the objective function $J(\omega_i)  = \mathbb{E} [R_i]$. This is done by adjusting the parameters $\omega_i$ in the direction of the gradient of $J(\omega_i)$ which can be written as:
	\begin{equation*}
	\nabla_{\omega_i} J(\omega_i) = \mathbb{E}_{s \sim \mathcal{D}} \big[ \nabla_{\omega_i} \bm\mu_{\omega_i}(\bm{o}_i) \nabla_{a_i} Q^{\bm\mu_{\omega_i}}(\bm{o}_i,a_i) |_{a_i=\bm\mu_\omega(\bm{o}_i)} \big] 
	\end{equation*}
	The actions $a$ are produced by the actor $\bm{\mu}_{\omega_i}$, are evaluated by the critic $ Q^{\bm\mu_i} $ which minimises the following loss:
	\begin{equation*}
	\mathcal{L}(\omega_i) = \mathbb{E}_{\bm{o}_i, a_i, r, \bm{o}'_i \sim \mathcal{D}} \Big[(Q^{\bm{\mu}_{\omega_i}}(\bm{o}_i, a_i) - y)^2 \Big]
	\end{equation*}
	where $\bm{o}'_i$ is the next observation, $\mathcal{D}$ is an experience replay buffer which contains tuples $(\bm{o}_i,\bm{o}'_i,a,r)$, $y =   r + \gamma Q^{\bm{\mu'}_{\omega}}(\bm{o}'_i, a'_i)$ represent the target Q-value. $Q^{\bm{\mu'}_{\omega_i}}$ is a target network whose parameters are periodically updated with the current parameters of $Q^{\bm{\mu}_{\omega_i}}$ to make training more stable. $\mathcal{L}(\omega_i)$ minimises the expectation of the difference between the current and the target action-state function.
	
	In this formulation, as there is no interaction between agents, the policies are learned independently. We adopt the CLDE paradigm by letting the critics $Q^{\bm{\mu}_{\omega_i}}$ use the observations $ \mathbf{x} = (\bm{o}_1, \bm{o}_2, \dots, \bm{o}_N)$ and the actions of all agents, hence: 
	\begin{equation} \label{eq:maddpg_a}
	\nabla_{\omega_i}J(\bm{\mu}_{\omega_i}) = \mathbb{E}_{\mathbf{x}, a \sim \mathcal{D}} \Big[\nabla_{\omega_i}\bm{\mu}_{\theta_i}(\bm{o}_i) \nabla_{a_i}Q^{\bm{\mu}_{\omega_i}}(\mathbf{x},a_1, a_2, \dots, a_N)|_{a_i=\bm{\mu}_{\omega_i}(\bm{o}_i)} \Big]
	\end{equation}
	where $ \mathcal{D} $ contains transitions in the form of  $ ( \mathbf{x}, \mathbf{x}', a_1, a_2, \dots, a_N, r_1, \dots, r_n )$ and $ \mathbf{x'} = (\bm{o}'_1, \bm{o}'_2, \dots, \bm{o}'_N) $ are the next observations of all agents. Accordingly, $ Q^{\bm{\mu}_{\omega_i}} $ is updated as
	\begin{equation}\label{eq:maddpg_q}
	\begin{aligned}
	\mathcal{L}(\omega_i) = {} & \mathbb{E}_{\mathbf{x}, a, r, \mathbf{x'} \sim \mathcal{D}} \Big[(Q^{\bm{\mu}_{\omega_i}}(\mathbf{x}, a_1, a_2, \dots, a_N) - y)^2 \Big], \\ 
	y = {} &  r_i + \gamma Q^{\bm{\mu'}_{\omega_i}}(\mathbf{x'}, a'_1, a'_2, \dots, a'_N)
	\}
	\end{aligned}
	\end{equation}
	in which $ a'_1, a'_2, \dots, a'_N $ are the next actions of all agents.
	By minimising Eq. \ref{eq:maddpg_q} the model attempts to improve the estimate of the critic $Q^{\bm{\mu}_{\omega_i}}$ which is used to improve the policy itselfs through Eq. \ref{eq:maddpg_a}.
	Since the input of the policy described in Eq. \ref{eq:policy} is $(\bm{o}_i, \mathbf{m})$ the gradient of the resulting algorithm to maximize $J(\theta_i)  = \mathbb{E} [R_i]$ can be written as:
	\begin{equation*}
	\nabla_{\theta_i}J(\bm{\mu}_{\theta_i}) =\mathbb{E}_{\mathbf{x}, a, \mathbf{m} \sim \mathcal{D}} \Big[\nabla_{\theta_i}\bm{\mu}_{\theta_i}(\bm{o}_i, \mathbf{m}) \nabla_{a_i}Q^{\bm{\mu}_{\theta_i}}(\mathbf{x}, a_1, \dots, a_N)|_{a_i=\bm{\mu}_{\theta_i}(\bm{o}_i, \mathbf{m})} \Big]
	\end{equation*}
	where $ \mathcal{D} $ is a replay buffer which contains transitions in the form of \\ $ ( \mathbf{x}, \mathbf{x}', a_1, \dots, a_N, \mathbf{m}, r_1, \dots, r_n )$. The $ Q^{\bm{\mu}_{\theta_i}} $ function is updated according to Eq. \ref{eq:maddpg_q}.
	Algorithm \ref{algo:MDMADDPG} provides the pseudo-code of the resulting algorithm, that we call MD-MADDPG (Memory-driven MADDPG).

	\subsection{MD-MADDPG decentralised execution}
	
	During execution, only the learned actors $\bm{\mu}_{\theta_1},\bm{\mu}_{\theta_2}, \dots, \bm{\mu}_{\theta_N} $ are used to make decisions and select actions. An action is taken in turn by a single agent. The current agent receives its private observations, $\bm{o}_i$, reads $\mathcal{M}$ to extract $\mathbf{r}_i$ (Eq. \ref{eq:readv}), generates the new version of $\mathbf{m}$ (Eq. \ref{eq:writev}), stores it into $\mathcal{M}$ and selects its action $a_i$ using $\bm{\mu}_i$. The policy of the next agent is then driven by the updated memory.

	\begin{algorithm}[h!]
		\caption{MD-MADDPG algorithm}\label{alg:CeMADDPG}
		\label{algo:MDMADDPG}
		{\color{black}
			\begin{algorithmic}[1]
				\State Inizialise actors ($ \bm{\mu}_{\theta_1}, \dots,  \bm{\mu}_{\theta_N} $) and critics networks ($ Q_{\theta_1}, \dots, Q_{\theta_N} $)
				\State Inizialise actor target networks ($ \bm{\mu'}_{\theta_1}, \dots,  \bm{\mu'}_{\theta_N} $) and critic target networks ($ Q'_{\theta_1}, \dots, Q'_{\theta_N} $)
				\State Inizialise replay buffer $ \mathcal{D} $
				\For{episode = 1 to E}
				\State Inizialise a random process $ \mathcal{N} $ for exploration
				\State Inizialise memory device $ \mathcal{M} $			
				\For{t = 1 to max episode length}
				\For{agent {\color{black}$i$} = 1 to $ N $}
				\State Receive observation $ \bm{o}_i $ and the message $ \mathbf{m} \leftarrow \mathcal{M} $
				\State Set $ \mathbf{m}_i = \mathbf{m} $
				\State Generate observation encoding $ \mathbf{e}_i $ (Eq. \ref{eq:encoding})
				\State Generate read vector $ \mathbf{r}_i $ (Eq. \ref{eq:readv})
				\State Generate new message $ \mathbf{m'} $ (Eq. \ref{eq:writev})
				\State Generate new time dependant noise istance $\mathcal{N}_t $ 
				\State Select action $  a_i = \varphi^{act}_{\theta_i}([\mathbf{e}_i,\mathbf{r}_i,\mathbf{m'}]) + \mathcal{N}_t $ 
				\State Store the new message in the memory device $ \mathcal{M} \leftarrow \mathbf{m'} $
				\EndFor		
				\State Set $ \mathbf{x} = (\bm{o}_1, \dots, \bm{o}_N) $ and $ \mathbf{\Phi} = ( \mathbf{m}_1,\dots,\mathbf{m}_N) $
				\State Execute actions $ \mathbf{a} = (a_1, \dots, a_N) $, observe rewards $ r $ and next observations $ \mathbf{x'} $
				\State Store ($ \mathbf{x},\mathbf{x'}, \mathbf{a},\mathbf{\Phi},r$) in replay buffer $ \mathcal{D} $		
				\EndFor
				\For{agent i = 1 to $ N $}
				
				\State Sample a random minibatch $ \Theta $ of $ B $ samples ($ \mathbf{x},\mathbf{x'}, \mathbf{a},\mathbf{\Phi},r$) from $ \mathcal{D} $
				\State Set $ y = r_i + \gamma Q^{\bm{\mu'}_{\theta_i}}(\mathbf{x'}, a'_1, \dots, a'_N) |_{a'_k =  \bm{\mu'}_{\theta_k}(o_k,\mathbf{m_k}) }  $
				\State Update critic by minimizing:\\
				\begin{equation*}
				\mathcal{L}(\theta_i) = \frac{1}{B} \sum_{( \mathbf{x},\mathbf{x'}, \mathbf{a},\mathbf{\Phi},r) \in  \Theta} (y -Q^{\bm{\mu}_{\theta_i}}(\mathbf{x}, a_1, \dots, a_N))^2
				\end{equation*}
				\State Update actor according to the policy gradient:
				\small
				\begin{equation*}
				\nabla_{\theta_i} J \approx \frac{1}{B} \sum_{( \mathbf{x},\mathbf{x'}, \mathbf{a},\mathbf{\Phi},r)}  \Big(\nabla_{\theta_i}\bm{\mu}_{\theta_i}(\bm{o}_i, \mathbf{m}_i) \nabla_{a_i}Q^{\bm{\mu}_{\theta_i}}(\mathbf{x}, a_1, \dots,a_i, \dots a_N)|_{a_i=\bm{\mu}_{\theta_i}(o_i,\mathbf{m}_i)} \Big)
				\end{equation*}
				\normalsize
				\EndFor
				\State Update target networks:
				\begin{equation*}
				\theta^{'}_i = \tau \theta_i + (1-\tau)\theta^{'}_i
				\end{equation*}					
				\EndFor
			\end{algorithmic}
		}
	\end{algorithm}

	\section{Experimental Settings and Results} \label{sec:experiments}
	
	\subsection{Environments} \label{sec:environments}
\begin{figure}[h]
		\subfloat[Cooperative Navigation]{\includegraphics[width = 0.2\linewidth]{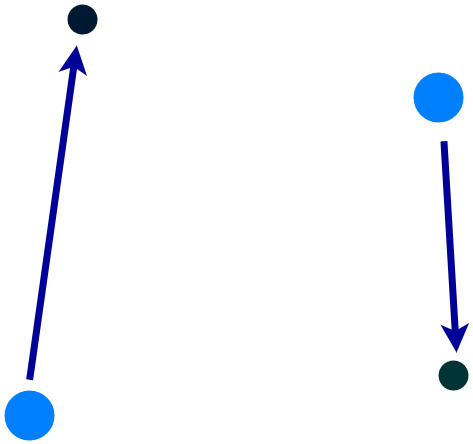}} 
		\hspace{1.5cm}
		\subfloat[Partial Observable CN]{\includegraphics[width = 0.2\linewidth]{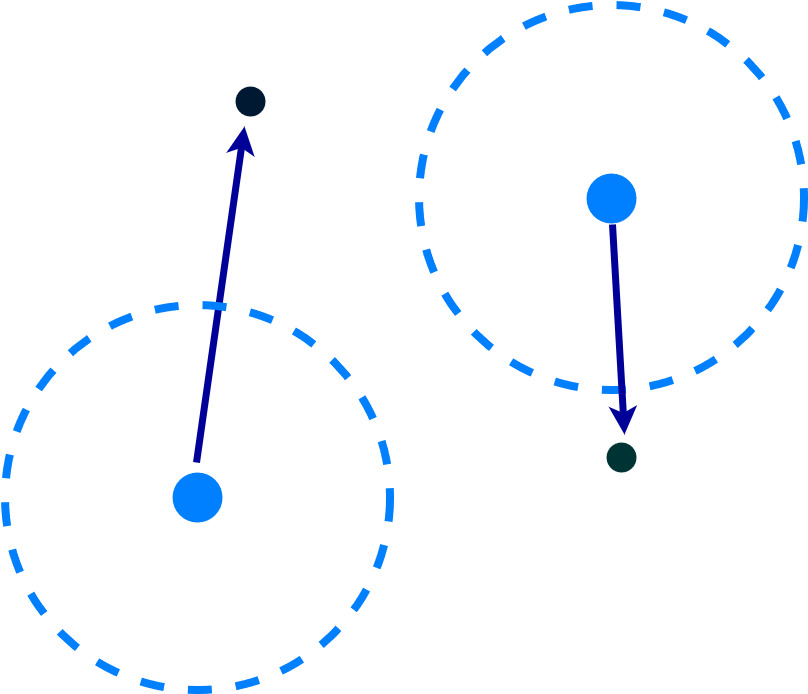}}
		\hspace{1.5cm}
		\subfloat[Synchronous CN]{\includegraphics[width = 0.2\linewidth]{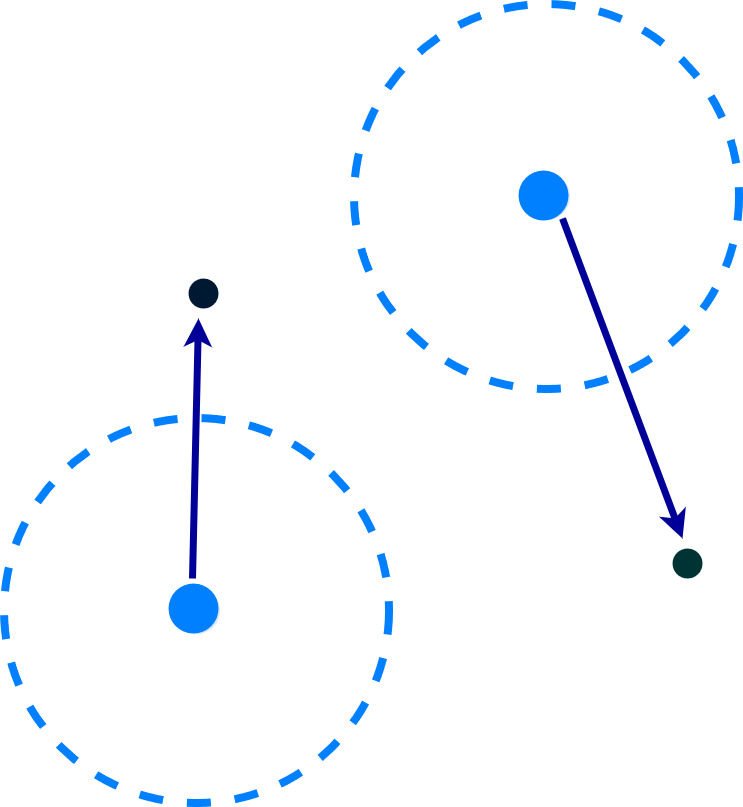}}\\
		\subfloat[Sequential CN]{\includegraphics[width = 0.2\linewidth]{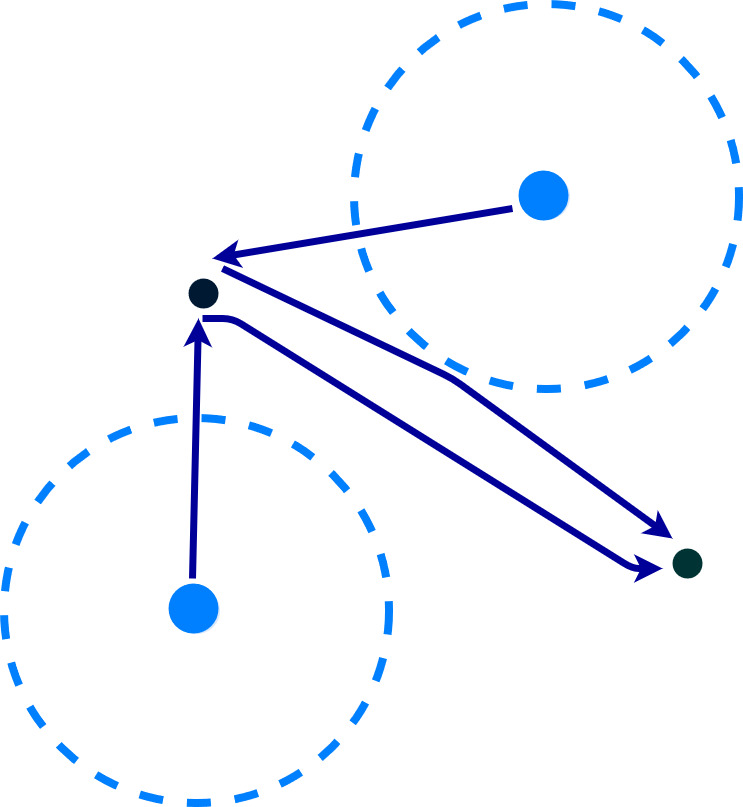}} 
		\hspace{1.5cm}
		\subfloat[Swapping CN]{\includegraphics[width = 0.2\linewidth]{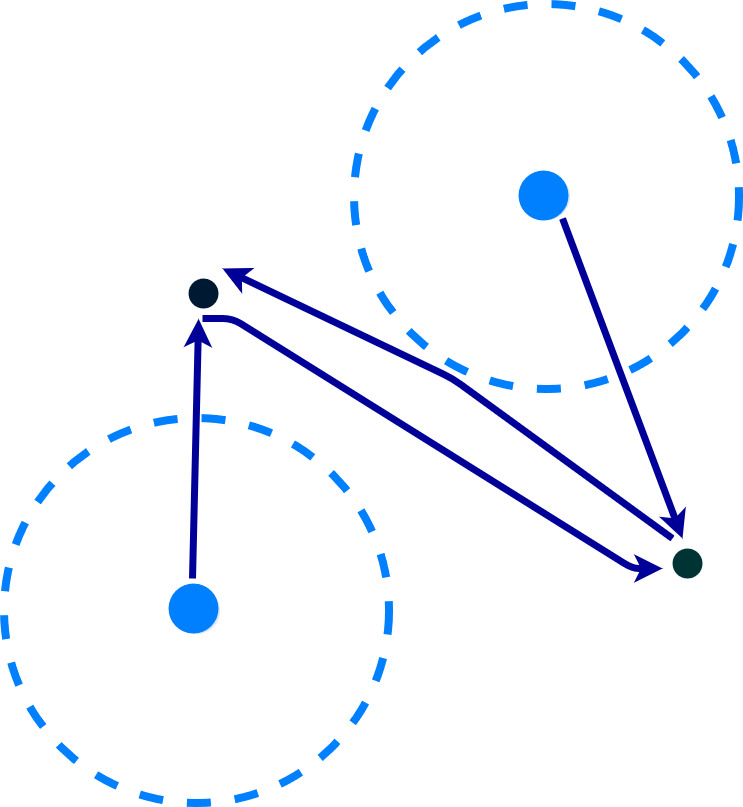}}
		\hspace{1.5cm}
		\subfloat[Waterworld]{\includegraphics[width = 0.2\linewidth]{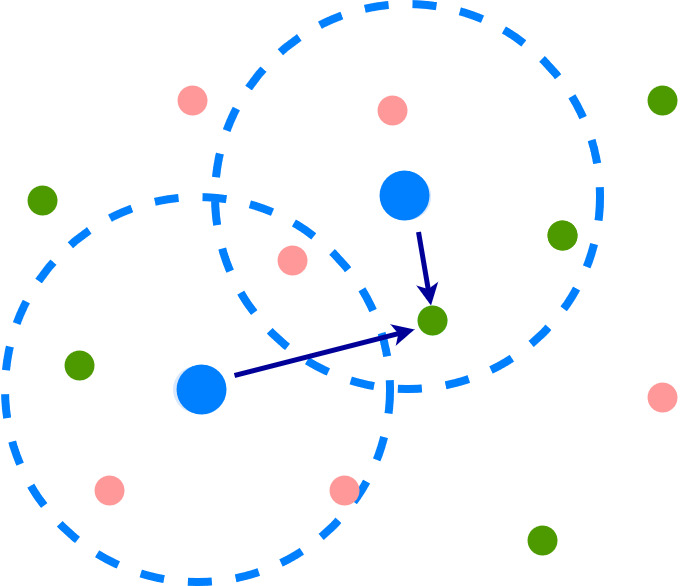}}
		\caption{An illustration of our environments. Blue circles represent the agents; dashed lines indicate the range of vision; green and red circles represent the food and poison targets, respectively, while black dots represent landmarks to be reached.}
		\label{fig:environments}
	\end{figure}

	In this section, we present a battery of six two-dimensional {\color{black}navigation environments} (Figure \ref{fig:environments}), with continuos space and discrete time. {\color{black}We introduce tasks of} increasing complexity, requiring progressively more elaborated coordination skills: five environments {\color{black} are inspired} by the Cooperative Navigation problem from the multi-agent particle environment \citep{lowe2017multi,mordatch2017emergence} {\color{black}in addition to} Waterworld from the SISL suite \citep{gupta2017cooperative}. We focus on two-agents systems to keep the settings sufficiently simple and {\color{black}attempt an initial analysis and interpretation of} emerging communication behaviours. A short description of the six environments is in order.
	
	\paragraph{Cooperative Navigation (CN).}
	This environment consists of $N$ agents and $N$ corresponding landmarks. An agent's task is to occupy one of the landmarks whilst avoiding collisions with other agents. 
	Every agent observes the distance to all others agents and landmark positions. 
	
	\paragraph{Partial Observable Cooperative Navigation (PO CN).}
	This is based on Cooperative Navigation, i.e. the task and action space are the same, but the agents now have a limited vision range and can only observe a portion of the environment around them within a pre-defined radius.
	
	\paragraph{Synchronous Cooperative Navigation (Sync CN).} 
	The agents need to occupy the landmarks exactly at the same time in order to be positively rewarded. A landmark is declared as occupied when an agent is arbitrarily close to it. Agents are penalised when the landmarks are not occupied at the same time.
	
	
	\paragraph{Sequential Cooperative Navigation (Sequential CN).}
	This environments is similar to the previous one, but the agents {\color{black}here need to occupy landmarks sequentially and avoid to reach them simultaneuosly} in order to be positively rewarded. Occupying the landmarks at the same time is penalised. 
	
	\paragraph{Swapping Cooperative Navigation (Swapping CN).} 
	In this case the task is more complex as it consists of two sub-tasks. Initially, the agents need to reach the landmarks and occupy them at same time. Then, they need to swap their landmarks and repeat the same process.
	
	\paragraph{Waterworld.} 
	In this environment, two agents with limited range vision have to collaboratively capture food targets whilst avoiding poison targets. A food target can be captured only if both agents reach it at the same time. Additional details are reported in \citep{gupta2017cooperative}.
	
	\subsection{Implementation Details}
	
	In all our experiments, we use a neural network with one layer {\color{black}(512 unites)} for the encoding (Eq. \ref{eq:encoding}), a neural network with one layer {\color{black}(256 units)} for the action selector (Eq. \ref{eq:actionselector}) and neural networks with three hidden layers {\color{black}($ 1024, 512, 256$ units, respectively)} for the critics. For MADDPG and MA-MADDPG the actors are implemented with neural networks with 2 hidden layers {\color{black}($ 512, 256$ units)}. 
	The size of the $\mathbf{m}$ is fixed to $ 200 $; {\color{black} this value that has been empirically found to be optimal given the network architectures (Section \ref{subsec:memorysizes} provides a validation study on the choice of memory size). Consequently, the size of $\mathbf{h_i} $ and $\mathbf{e}_i$ is set to $200$.}
	{\color{black}We use the} Adam optimizer \citep{kingma2014adam} with a learning rate of $ 10^{-3} $ for critic and $ 10^{-4} $ for policies. The reward discount factor is set to $ 0.95 $, the size of the replay buffer to $ 10^{6} $ and the batch size to $ 1,024 $. The number of time steps for episode is set to $1,000$ for Waterworld and $100$ for the other environments. We update network parameters after every 100 samples added to the replay buffer using soft updates with $ \tau = 0.01 $. 
	{\color{black} We train all the models over $ 60,000 $ episodes of 100 time-steps each on all the environments, except for Waterworld for which we use $ 20,000 $ episodes of $ 1,000 $ time-steps each for training.}
	{\color{black} The Ornstein-Uhlenbeck process \citep{uhlenbeck1930theory} with $ \theta = 0.15 $ and $ \sigma =  0.3 $ is a stochastic process which, over time, tends to drift towards its mean. This is commonly employed within DDPG \citep{lillicrapHPHETS15} and in order to introduce temporally correlated noise. Doing so it is possible to avoid the effects of averaging random decorrelated signals which would lead a less effective exploration. Discrete actions are supported by the Gumbel-Softmax, a biased, low-variance gradient estimator \citep{jang2016categorical}. This estimator is typically used within the back-propagation algorithm in the presence of categorical variables. 
		We use Python 3.5.4 \citep{van1995python} with PyTorch v0.3.0 \citep{paszke2017automatic} as software for automatic differentiable computing and machine learning framework. All the computations were performed using Intel(R) Xeon(R) CPU E5-2650 v3 @ 2.30GHz as CPU and GeForce GTX TITAN X as GPU.
	}
	
	\subsection{Experimental results}
	
	In our experiments, we compared the proposed {\color{black} MD-MADDPG against four algorithms: MADDPG \citep{lowe2017multi}, Meta-agent MADDPG (MA-MADDPG), CommNet \citep{sukhbaatar2016learning} and MAAC \citep{iqbal2018actor}}. 
	{\color{black}MA-MADDPG is a variation of MADDPG in which} the policy of an agent during both training and execution is conditioned upon the observations of all the other agents in order to overcome difficulties due to partial observability. 
	{\color{black}These methods have been selected to provide fair comparisons since they offer different learning approaches in multi-agent problems. MADDPG is what our method builds on so this comparison can quantify the improvements brought by the proposed communication mechanism; MA-MADDPG offers an alternative information sharing mechanism; CommNet implements an explicit form of communication; MAAC is a recent is a state-of-the-art method in which critics select information to share through an attention mechanism.}
	We analyse the performance of these competing learning algorithms on all the six environments described in Section \ref{sec:environments}. In each case, after training, we evaluate an algorithm's performance by collecting samples from an additional $1,000$ episodes, which are then used to extract different performance metrics: the {\it reward} quantifies how well a task has been solved; the {\it distance from landmark{\color{black}s}} captures how {\color{black}closely} an agent has reach{\color{black}ed the landmarks}; the {\it number of collisions} counts how many times an agent has failed to avoid collisions with others;  {\it sync occupations} counts how many times {\color{black}the} landmarks have been occupied simultaneously and, analogously, {\it not sync occupations} counts how many times only one of the two landmarks has been occupied. 
	For Waterworld, we {\color{black}also} count the {\it number of food targets} and {\it number of poison targets}. {\color{black} Since this environment requires continuous actions, we cannot use MAAC as this method only operates on discrete action spaces.}  In Table  \ref{tab:all_results}, for each metric, we report the sample average and standard deviation obtained by each algorithm on each environment. {\color{black} A visualization of all results through boxplot can be found in Section \ref{subsec:mainresultsvisualization} in Supplementary Material. Illustrative videos to show the performance of MD-MADDPG on the environments are publicly available \footnote{Supplementary illustrative videos: \url{https://www.youtube.com/watch?v=P9XWdpmsEy8}}.
		
}
	
	\begin{table*}[h]
		\begin{center}
			\scalebox{0.9}{
				{\color{black}
					\begin{tabular}{lllllll}
						\textbf{Environment}		&  \textbf{Metric} & \textbf{MADDPG} & \textbf{MA-MADDPG} & \textbf{CommNet} & \textbf{MAAC} &\textbf{MD-MADDPG}
						\\ \hline 
						&     Reward     &   $ -2.30 \pm 0.11 $      & $ -2.29 \pm 0.10 $&  $ -2.7 \pm 0.26 $ & $  -4.72 \pm 1.35$ &  $ \bm{-2.27 \pm 0.10}  $ \\ 
						CN	&    Average distance     &   $0.15 \pm 0.051   $     & $ 0.14 \pm 0.05 $  &  $ -0.35 \pm 0.13 $ &  $ 1.36 \pm 0.67 $  &    $ \bm{0.13  \pm 0.05}$\\
						&	  \# collisions & $ \bm{0.11 \pm 0.76} $  & $ 0.17  \pm 0.90$ & $ 0.19 \pm 1.06 $ & $ { 0.58 \pm 1.42 }$ & $ 0.12     \pm 0.82 $
						\\ \hline 
						&     Reward     &   $ \bm{-2.62  \pm  0.34}  $    &  $ -2.67  \pm  0.38 $ & $ -2.78 \pm 0.43  $  & $ -3.17 \pm 0.62 $ &    $ -2.68  \pm 0.46  $\\ 
						PO CN  	&    Average distance       &    $ \bm{0.30 \pm 0.17}  $    & $ 0.33 \pm 0.19 $ & $  0.39\pm0.21  $ & $ 1.26 \pm 2.53  $&  $0.34 \pm 0.22 $ \\     
						&	  \# collisions & $ 0.55 \pm 1.64 $  & $ \bm{0.14 \pm 0.69} $   & $ 0.37 \pm 1.48  $&$ 0.58 \pm 0.31 $& $ 0.26  \pm 1.06 $
						\\ \hline
						&     Reward     &  $ 75.83 \pm 72.23    $   & $ \bm{192.92 \pm 29.78} $  &$  188.02 \pm35.41  $  & $  161.35 \pm 80.03  $ &      $ 92.90 \pm 69.78  $\\ 
						Sync CN	&  \# sync occup.       &   $ 26.71 \pm 19.86	 $        & $ \bm{53.96 \pm 20.16} $ & $3.62 \pm12.14  $ & $ 139.56 \pm 63.55  $&     $ 31.6 \pm 19.34 $ \\ 
						&	  \# not sync occup. &  $ 21.36 \pm 16.60  $   & $41.75 \pm 56.25 $ & $  	46.35 \pm 29.47  $ &$  44.84 \pm 58.71   $ & $ \bm{17.58 \pm 12.00}  $
						\\ \hline
						Sequential CN &     Reward     &   $ 125.98 \pm 33.4  $     &   $117.52  \pm 35.62$  &$  131.67 \pm19.48  $   &$ 90.11 \pm21.33  $  &   $ \bm{130.15  \pm 35.19}  $ \\ 
						&     Average distance       &   $  260.16  \pm 14.41	 $        &     $114.7 \pm 45.71$ &$ 102.63 \pm34.96 $   &$ 101.11 \pm40.71  $  &   $ \bm{99.15 \pm 50.59} $
						\\ \hline
						Swapping CN &     Reward     & $ 125.60 \pm 50.13     $        &  $86.99 \pm 68.52$ &$  109.55 \pm56.64  $   &$  75.71\pm69.80  $  &      $ \bm{129.63 \pm 47.26 } $\\ 
						&     Average distance       &    $ 76.70 \pm 30.24	 $        &  $132.77  \pm 89.98$ & $  123.54 \pm84.9  $  &$  152.7\pm43.72  $  &   $ \bm{53.21  \pm 40.80} $
						\\ \hline
						&     Reward     &   $ 262.29 \pm 141.07  $     & $99.31 \pm 118.31$ &$  139.29 \pm121.42 $   &$  NA  $  &    $ \bm{503.96} \pm 103.91  $ \\ 
						Waterworld	&    \# food targets            &  $ 13.91 \pm 7.30 $ & $5.25 \pm 6.07$ &$  	10.2 \pm7.1  $   &$  NA  $  &   $ \bm{26.25 \pm 5.41}  $  \\
						&	  \# poison targets & $ 8.61 \pm 3.32$  & $ \bm{5.34 \pm 2.45} $ & $  8.01 \pm5.22  $  &$  NA  $  & $ 7.77 \pm 2.95 $
						\\ \hline
					\end{tabular}
				}
			}
			\caption{ {\color{black} Comparison of MADDPG, MA-MADDPG, CommNet, MAAC and MD-MADDPG }on six environments ordered by increasing level of difficulty, from CN to Waterword. The sample mean and standard deviation for $1,000$ episodes are reported for each metric.}
			\label{tab:all_results}
		\end{center}
	\end{table*}
	All algorithms perform very similarly in the Cooperative Navigation and Partial Observable Navigation cases. This result is expected because these environments involve {\color{black} relatively} simple tasks that can be completed {\color{black}even} without explicit message-passing and information sharing functionalities. Despite communication not being essential, MD-MADDPG reaches comparable performance to MADDPG and MA-MADDPG. In the Synchronous Cooperative Navigation case, the ability of MA-MADDPG to overcome partial observability issues by sharing the observations across agents seem to be crucial as the total rewards achieved by this algorithm are substantially higher than those obtained by both MADDPG and MD-MADDPG. In this case, whilst not achieving the highest reward, MD-MADDPG keeps the number of unsynchronised occupations at the lowest level, and also performs better than MADDPG on all three metrics. It would appear that in this case pulling all the private observations together is sufficient for the agent{\color{black}s} to synchronize their paths leading to the landmarks.  
	
	When moving on to more complex tasks requiring further coordination, the performances of the three algorithms diverge further in favour of MD-MADDPG. The requirement for strong collaborative behaviour is more evident in the Sequential Cooperative Navigation problem as the agents  need to explicitly learn to take either shorter or longer paths from their initial positions to the landmarks in order to occupy them in sequential order.  Furthermore, according to the results in Table \ref{tab:all_results}, the average distance travelled by the agents trained with MD-MADDPG is less then the half the distance travelled by agents trained with MADDPG, indicating that these agents were able to find a better strategy by developing an appropriate communication protocol. Similarly, in the Swapping Cooperative Navigation scenario, MD-MADDPG achieves superior performance, and is again able to discover solutions involving the shortest f. Waterworld is significantly more challenging as it requires a sustained level of synchronization throughout the entire episode and can be seen as a sequence of sub-tasks whereby each time the agents must reach a new food target whilst avoiding poison targets. In Table \ref{tab:all_results}, it can be noticed that MD-MADDPG significantly outperforms both competitors in this case. The importance of sharing observations with other agents can also be seen here as MA-MADDPG generates good policies that avoid poison targets, yet in this case, the average reward is substantially lower than the one scored by MD-MADDPG. 
	{\color{black} Experimental settings so far have involved two agents. In addition, we have also investigated settings with a higher number of agents, see Supplementary Material (Section \ref{subsec:moreagents_CN} for Cooperative Navigation and Section \ref{subsec:moreagents_POCN} for Partial Observable Cooperative Navigation). These results show, that the prososed method can be successfully used on larger systems without incurring any numerical complications or convergence difficulties. When comparing to other algorithms, MD-MADDPG has resulted in superior performance performance are indeed achieved on Cooperative Navigation with respect to the reward metric. On Partially Observable Cooperative Navigation, there is no definite winner, nevertheless MD-MADDPG shows competitive performance, for example it outperforms all the baselines on the 5 agents scenario.}
	
	{\color{black} In Supplementary Material in Section \ref{subsec:ablatation_context}, we provide an ablation study showing that the main components of MD-MADDPG are needed for its correct behaviour. We investigate the effects of removing either one of the key components, i.e. context vector, read and write modules. Removing the context vector reduces the quality of the performance obtained on CN and on environments which require greater coordination efforts, like Sequential CN, Swapping CN and Waterworld. On PO-CN no significant differences in performance are reported, while on Synchronous CN sync occupations worsen (of approximatively five times the amount) and sync occupations improve (of approximatively twice the amount). This result is explained by the fact that in Sync CN, good strategies that do not involve explicit communication can be learnt to achieve good performance on sync occupations. The best overall performance method on this scenario is MA-MADDPG (see Table \ref{tab:all_results}). This comparative method implements an implicit form of communication that is equivalent to a simple information sharing which can be very effective to overcome the partial observability issue which is the main challenge in Sync CN.
		We have observed that without the writing or reading components the performance worsened on all the run experiments.}
	
	\begin{figure*}[h] 
		\begin{minipage}[b]{0.45\linewidth}
			\centering
			\includegraphics[width=1\linewidth]{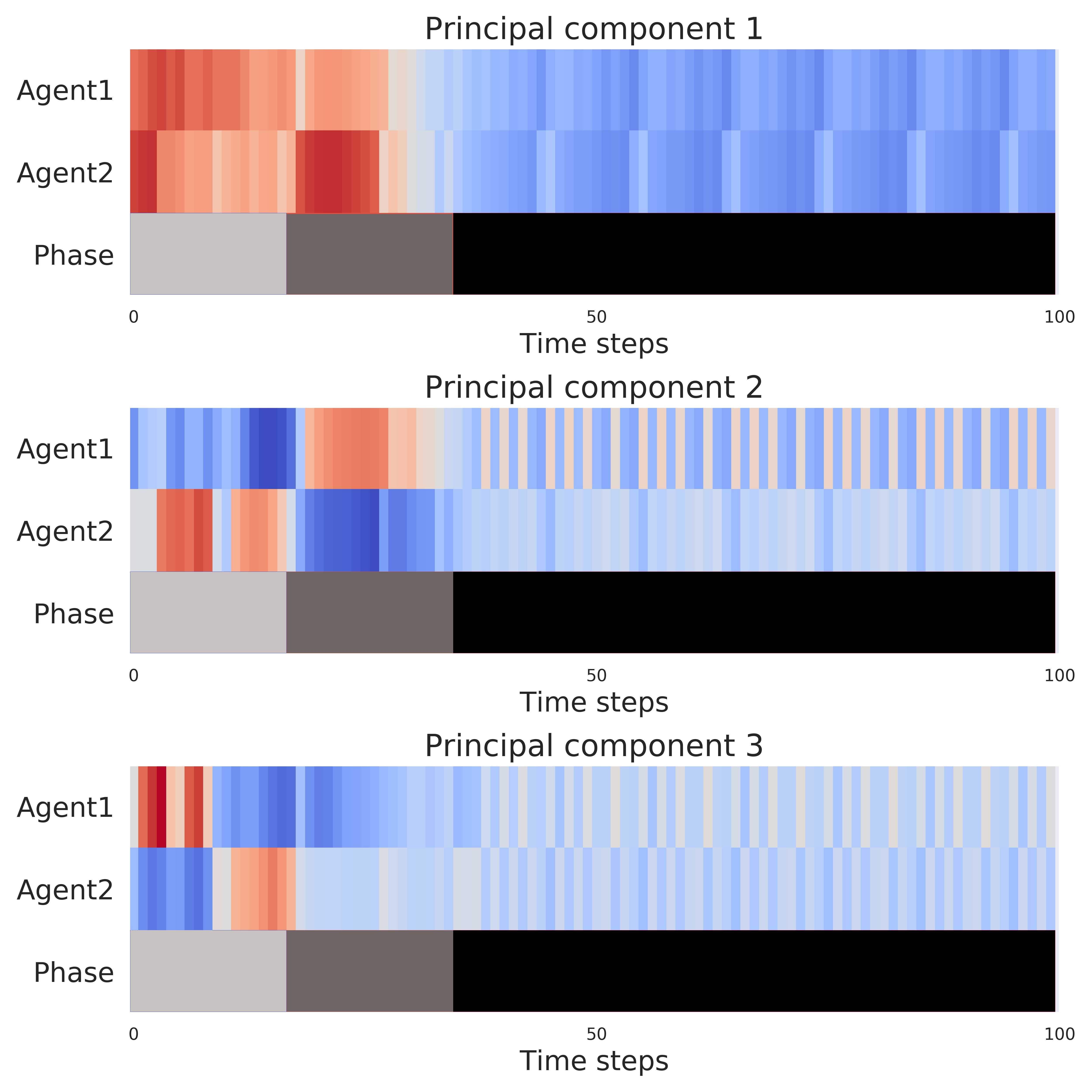} 
			\\ (a) Sequential CN
			\vspace{1ex}
		\end{minipage}
		\begin{minipage}[b]{0.45 \linewidth}
			\centering
			\includegraphics[width=1\linewidth]{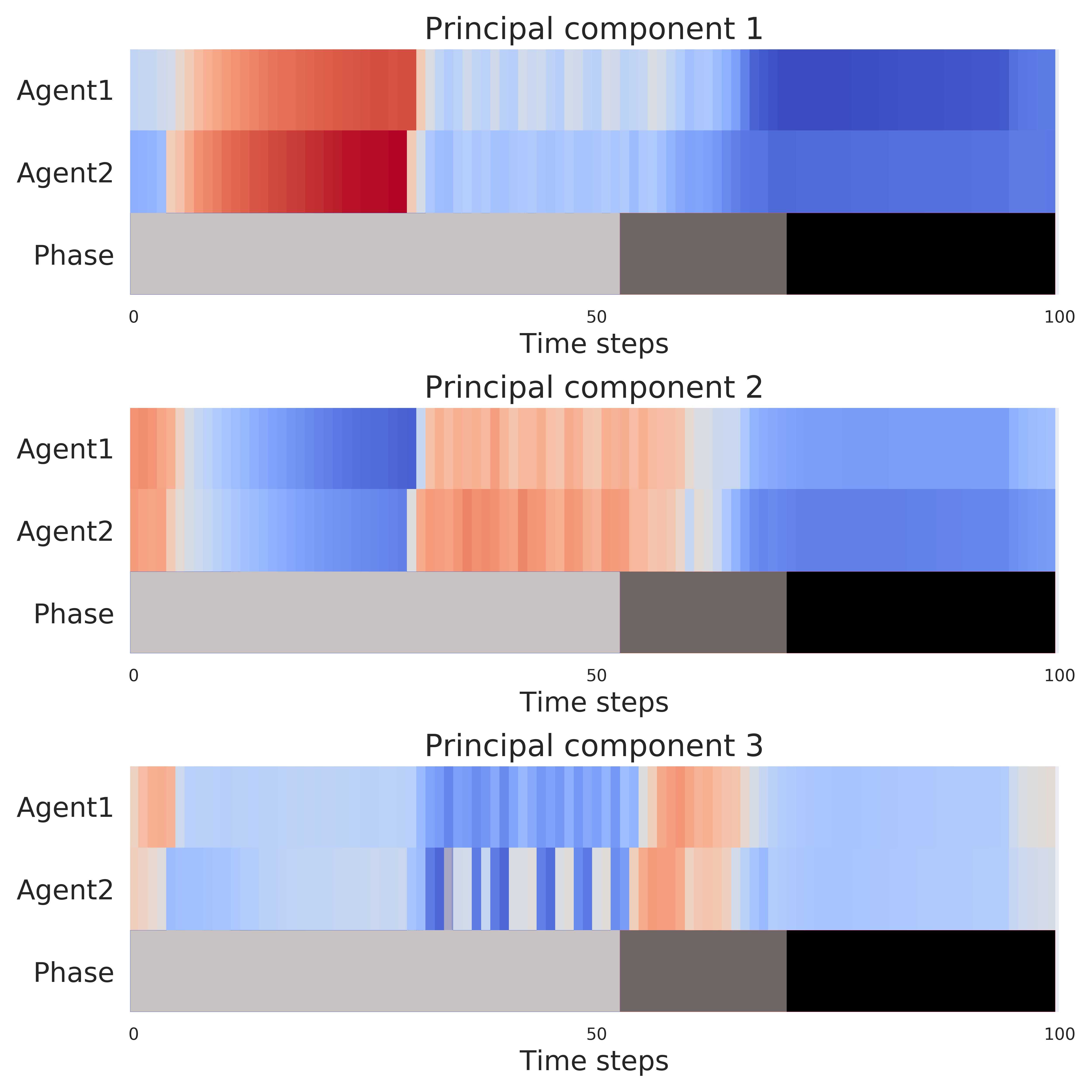} 
			\\ (b) Swapping CN
			\vspace{1ex}
		\end{minipage} 
		\\
		\begin{minipage}[b]{0.45\linewidth}
			\centering
			\includegraphics[width=1\linewidth]{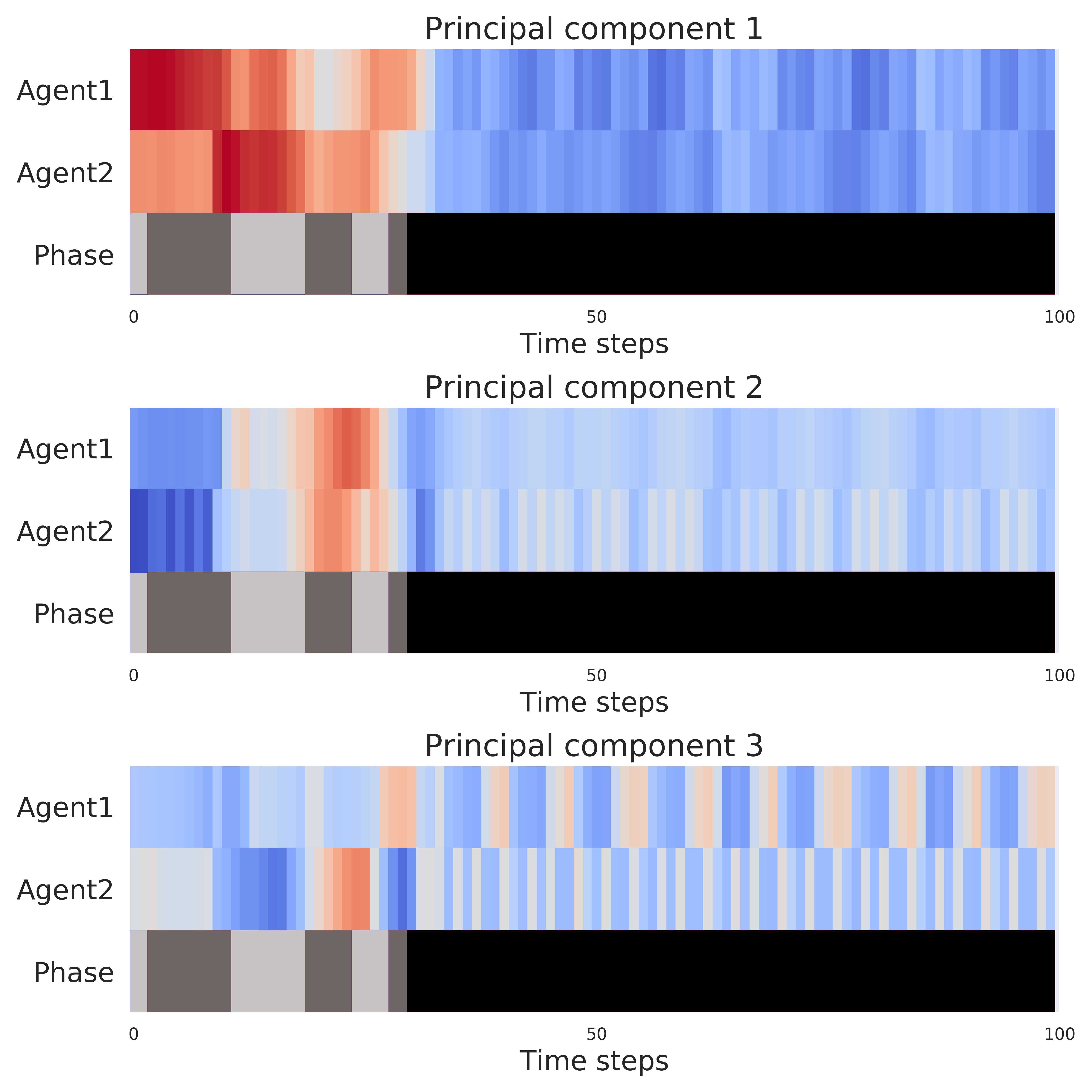} 
			\\ (c) Synchronous CN
			\vspace{1ex}
		\end{minipage} 
		\begin{minipage}[b]{0.45\linewidth}
			\centering
			\includegraphics[width=1\linewidth]{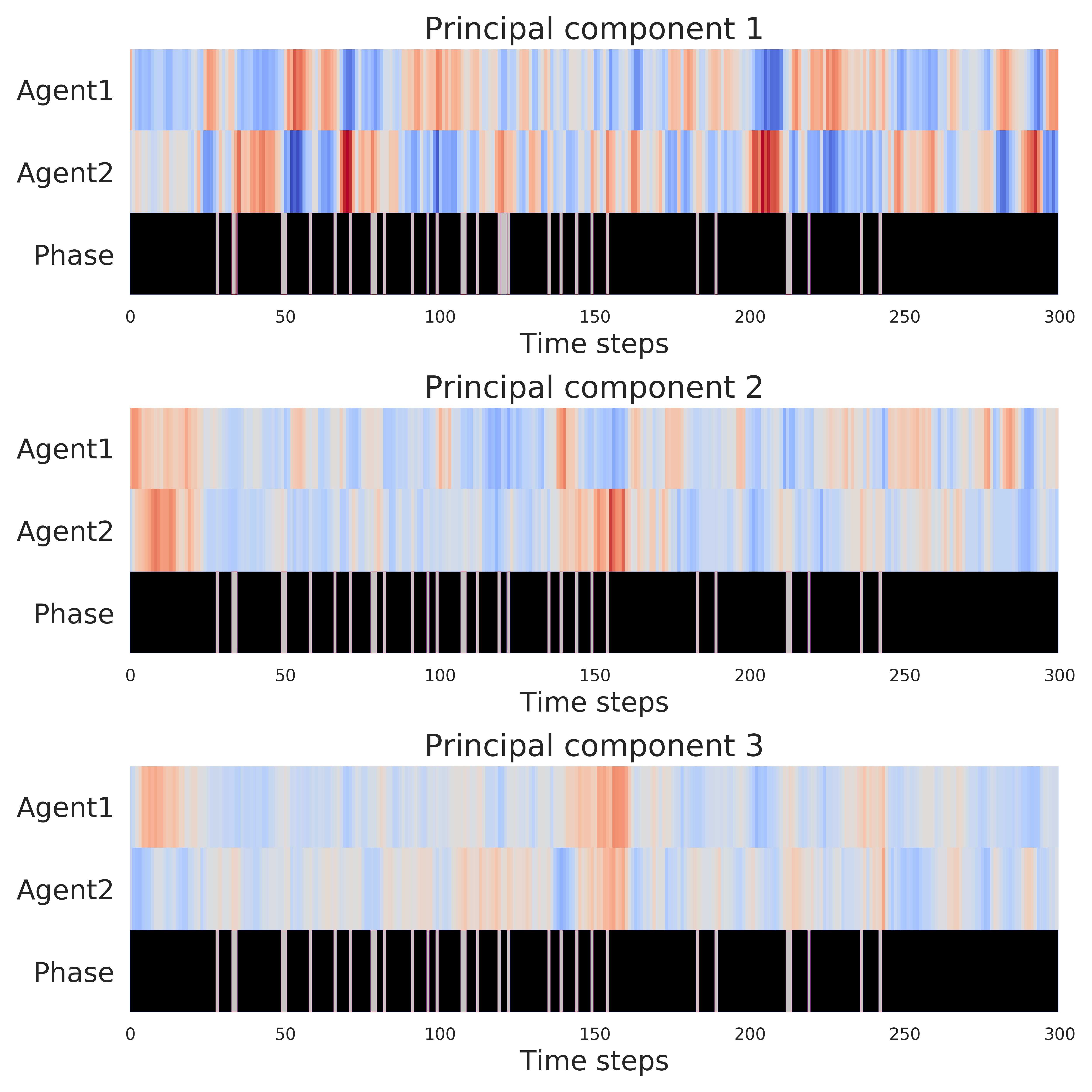} 
			\\ (d) Waterworld
			\vspace{1ex}
		\end{minipage}
		
		\caption{Visualisation of communications strategies learned by the agents in four different environments: the three principal components provide orthogonal descriptors of the memory content written by the agents and are being plotted as a function of time. Within each component, the highest values are in red, and the lowest values are in blue. The bar at the bottom of each figure indicates which phase (or sub-task) was being executed within an episode; see Section \ref{sec:commanalysis} for further details. The memory usage patterns learned by the agents are correlated with the underlying phases and the memory is no longer utilised once a task is about to be completed. 
		}
		\label{fig:memory_PC3} 
	\end{figure*}

	\subsection{Communication analysis} \label{sec:commanalysis}
	
	In this section, we explore the dynamic patterns of communication activity that emerged in the environments presented in the previous section, and look at how the agents use the shared memory throughout an episode while solving the required task. For each environment, after training, we executed episodes with time horizon $T$ and stored the write vector $\mathbf{m'}$ of each agent at every time step $t$. Exploring how $\mathbf{m'}$ evolves within an episode can shed some light onto the role of the memory device at each phase of the task. 
	{\color{black}The analysis presented in this section focuses on the write vector as we expect it to be stronger correlated with the environment dynamics than the other components. The content of the writing vector corresponds to the content of the communication channel itself, and is expected to contain information related to the task (e.g. changes in current environment, agent's strategy or observed point of interests). A communication analysis with respect to the read vector $\mathbf{r}_i$ is presented in Supplementary Material (Section \ref{subsec:readvector}). The content of the reading vector is an implicit representation internal to the agent itself that serves to interpret the content of the channel and at the same time to be utilised in the generation of $\mathbf{m'}$.} 
	In order to produce meaningful visualisations, we first projected the dimensions of $\mathbf{m'}$ onto the directions maximising the sample variance (i.e. the variance of the observed $\mathbf{m'}$ across simulated episodes) using a linear PCA. 
	
	Figure \ref{fig:memory_PC3} shows the principal components (PCs) associated with the two agents over time for four of our six simulation environments. Only the first three PCs were retained as those were found to cumulatively explain over $80\%$ of variance in all cases. The values of each PC were standardised to lie in $[0,1]$ {\color{black}in order have them in the same in same range for fair comparisons} and are plotted on a color map: one is in red and zero in blue. The timeline at the bottom of each figure indicates which specific phase of an episode is being executed at any given time point, and each consecutive phase is coloured using a different shade of grey. For instance, in Sequential Cooperative Navigation, a single landmark is reached and occupied in each phase. In Swapping Cooperative Navigation, during the first phase the agents search and find the landmarks; in the second phase they swap targets, and in the third phase they complete the task by reaching the landmarks again. In the Synchronous Cooperative Navigation the phase indicates if none of the landmarks is occupied (light-grey), if just one is occupied (dark-grey) and if both are occupied (black). Usually, in the last phase, the agents learn to stay close to their targets.  
	{\color{black} This analysis pointed out that in the final phases, when tasks are already completed and there is no need of coordination, the PCs representing the communication activities assume lower (blue values), while during previous phases, when tasks are still to be solved and cooperation is stronger required, they assume higher values (red).
		This led us to} interpret the higher values as being indicative of high memory usage, and lower values as being associated to low activity. In most cases, high communication activity is maintained when the agents are actively working and completing a task, while during the final phases (where typically there is no exploration because the task is considered completed) low activity levels are more predominant. 
	
	This analysis also highlights the fact that the communication channel is used differently in each environment. In some cases, the levels of activity alternate between agents. For instance, in Sequential Cooperative Navigation (Figure \ref{fig:memory_PC3}a), high levels of memory usage by one agent are associated with low ones by the other. A different behaviour is observed for the other environments, indeed in Swapping Cooperative Navigation task where both agents produce either high or low activation value, whereas in Synchronous Cooperative Navigation the memory activity is very intense before the phase three, while agents are collaborating to complete the task. The dynamics characterizing the memory usage also change based on the particular phase reached within an episode. For example, in Figure \ref{fig:memory_PC3}a, during the first two phases the agents typically show alternating activity levels whilst in the third phase both agents significantly decrease their memory activity as the task has already been solved and there are no more changes in the environment. Figure \ref{fig:memory_PC3} provides some evidence that, in some cases, a peer-to-peer communication strategy is likely to emerge instead of a master-slave one where one agent takes complete control of the shared channel. The scenario is significantly more complex in Waterworld where the changes in memory usage appear at a much higher frequency due to the presence of very many sequential sub-tasks. Here, each light-grey phase indicates that a food target has been captured. Peaks of memory activity seem to follow those events as the agents reassess their situation and require higher coordination to jointly decide what the next target is going to be.
	In Supplementary Material (\ref{subsec:corrupting}) we provide further experimental results showing the importance of the communication by corrupting the memory content at execution time, which further corroborate the role of the exchanged messages in improving agents' coordination.

	\section{Conclusions}\label{sec:conclusions}
	
	In this work, we have introduced MD-MADDPG, a multi-agent reinforcement learning framework that uses a shared memory device as an intra-agent communication channel to improve coordination skills. The memory content contains a learned representation of the environment that is used to better inform the individual policies. The memory device is learnable end-to-end without particular  constraints other than its size, and each agent develops the ability to modify and interpret it. We empirically demonstrated that this approach leads to better performance in {small-scale} {\color{black}(up to 6 agents in our experiments)} cooperative tasks where coordination and synchronization are crucial to a successful completion of the task and where world visibility is very limited. Furthermore, we have visualised and analyzed the dynamics of the communication patterns that have emerged in several environments. This exploration has indicated that, as expected, the agents have learned different communication protocols depending upon the complexity of the task. In this study we have mostly focused on two-agent systems to keep the settings sufficiently simple to understand the role of the memory.
	{Very competitive results have been obtained when more agents are used.
	
		In future work, we plan on studying the role played by the sequential order in which the memory is updated, as the number of agents grows.}
	A possible approach may consist of deploying agent selection mechanisms, possibly based on attention, so that only a relevant subset of agents can modify the memory at any given time, or impose master-slave architectures. 
	{A possible solution would be to have an agent acting as \textquotedblleft scheduler\textquotedblright  that controls the access to the memory, decides which information can be shared and provides scheduling for the writing accesses. Introducing such a scheduling agent would allow to keep the current framework unaltered, e.g. the sequential access to the memory would be retained. Although the scheduling agent would add an additional layer of complexity, this might reduce the number of memory access required in larger scale systems and improve the overall scalability.} In future work{, we will also apply} MD-MADDPG on environments characterized by more structured and high-dimensional observations (e.g. pixel data) where collectively learning to represent the environment through a shared memory should be particularly beneficial. 
	

	\bibliographystyle{plainnat}
	\bibliography{refs}


	\newpage
	\appendix
	\section*{Supplementary Material} \label{sec:supplementary}
	
	\section{Meta-agent}\label{sec:metaagent}

	In the meta-agent MADDPG (MA-MADDPG) each agent can observe the observations of all other agents to address issues related to partial observability. The policy of each agent is defined as $ \bm{\mu}'_i = \mathcal{O}_1, \times \mathcal{O}_2 \times \dots \times \mathcal{O}_N$ and the  gradient is:
	\begin{equation*}
	\begin{aligned}
	\nabla_{\theta_i}J(\bm{\mu}_{\theta_i}) = {} & \\
	{} &  \hspace{-1.5cm} \mathbb{E}_{\mathbf{x}, a \sim \mathcal{D}} \Big[\nabla_{\theta_i}\bm{\mu}_{\theta_i}(\mathbf{x}) \nabla_{a_i}Q^{\bm{\mu}_{\theta_i}}(\mathbf{x}, a_1, \dots, a_N)|_{a_i=\bm{\mu}_{\theta_i}(\mathbf{x})} \Big].
	\end{aligned}
	\end{equation*}
	where $\mathbf{x} = \bm{o}_1,\bm{o}_2, \dots, \bm{o}_N$  and $Q^{\bm{\mu}_{\theta_i}}$ is updated according to \ref{eq:maddpg_q}.

	\section{Additional Experiments} \label{sec:additional_exp}
	\subsection{Corrupting the memory}\label{subsec:corrupting}
	\renewcommand{\thefigure}{B\arabic{figure}}
	\setcounter{figure}{0}	
	\renewcommand{\thetable}{B\arabic{table}}
	\setcounter{table}{0}		
	Table \ref{tab:noise_results} shows the performance of MD-MADDPG when a Gaussian noise (mean 0 and standard deviation 1) is added to the memory content $\mathbf{m}$ at execution time. It can be noted that the corruption of the communication channel causes a general worsening of the performance across metrics.
	This shows that the messages exchanged by the agents are crucial to {\color{black}achieving} good performance, {\color{black}and that corrupting the memory hinders the communication which has a negative effect on synchronization}.
	
	\begin{table*}[h]
		\begin{center}
			
			\begin{tabular}{lllll}
				\textbf{Environment}		&  \textbf{Metric} & \textbf{MD-MADDPG - noise}
				\\ \hline 
				&     Reward     &      $ -2.28 \pm 0.1 $ \\ 
				CN	&    Average distance     &      $ 0.15 \pm 0.051 $ \\
				&	  \# collisions  & $0.11 \pm 0.76 $&
				\\ \hline 
				&     Reward     &      $ -2.68 \pm 0.45 $ \\ 
				PO-CN	&    Average distance     &      $ 0.34 \pm 0.22 $\\
				&	  \# collisions  & $0.32 \pm 1.14  $
				\\ \hline
				&     Reward       &      $ 187.17 \pm 41.84  $  \\ 
				Sync CN	&  \# sync occup.        &     $33.27 \pm 39.07 $  \\ 
				&	  \# not sync occup.  & $ 102.61 \pm 41.43 $ 
				\\ \hline
				Sequential CN 
				&     Reward       &   $ 124.72 \pm 30.28  $  \\ 
				&     Average distance        &   $ 111.27 \pm 52.66 $ 
				\\ \hline
				Swapping CN &     Reward      &      $124.93 \pm 44.48 $  \\ 
				&     Average distance        &   $ 112.44 \pm 83.88 $  
				\\ \hline
				&     Reward     &    $ 24.07 \pm 26.61  $  \\ 
				Waterworld	&    \#  food targets            &   $ 1.65 \pm 1.33  $   \\
				&	  \# poison targets  & $  10.37 \pm 3.91 $  
				\\ \hline
			\end{tabular}
			\caption{{\color{black}Performance of MD-MADDPG when Gaussian noise is added to the memory content at test time.}}
			\label{tab:noise_results}
			
		\end{center}
	\end{table*}

	\clearpage
	\subsection{Increasing the number of agents - Cooperative Navigation} \label{subsec:moreagents_CN}
	{\color{black}
		Table \ref{tab:exp_cn_moreagents} shows the comparison of MADDPG, MA-MADDPG, CommNet, MAAC and MD-MADDPG on Cooperative Navigation when the number of agents increases. 
		MD-MADDPG have the best performance achieving the highest reward on all the scenarios. MAAC shows higher performance in collision avoidance and CommNet in distance travelled (five and six agents).
		
		\begin{table*}[h!]
			\begin{center}
				{\color{black}
					\begin{tabular}{lllllll}
						\textbf{\# agents}	 &  \textbf{Metric}	&  \textbf{MADDPG} & \textbf{MA-MADDPG} & \textbf{CommNet} & \textbf{MAAC} & \textbf{MD-MADDPG}
						\\ \hline 
						&     Reward     &   $ -4.02 \pm 0.32 $      & $ -4.03 \pm0.29 $ & $ -4.66 \pm0.35 $ & $ -7.38 \pm1.28 $ & $ \mathbf{-3.75 \pm 0.21} $ \\ 
						3	&    Average distance     &   $ 0.34  \pm 0.1  $ & $ 0.34 \pm0.09 $ & $0.53 \pm0.11  $ & $ 1.45 \pm0.43 $    & $\mathbf{ 0.25 \pm 0.07} $  \\
						&	  \# collisions & $ 1.24  \pm 3.76   $  & $ 1.18 \pm2.2 $ & $ 5.94 \pm11.0 $ & $ 2.95 \pm5.04 $   & $\mathbf{ 1.15 \pm 2.34 }$  
						\\ \hline 
						&     Reward     &   $-6.86 \pm 0.68 $      & $ -7.0 \pm0.77 $ & $ -7.47 \pm0.64 $ & $ -12.82 \pm1.87 $ & $ \mathbf{-6.12 \pm 0.52} $ \\ 
						4	&    Average distance     &   $ 0.7 \pm 0.17   $    & $ 0.73 \pm0.19 $ & $ 0.81 \pm0.18 $ & $ 2.19 \pm0.47 $ & $ \mathbf{0.51 \pm 0.12} $  \\
						&	  \# collisions & $ 7.44 \pm 14.82 $  & $ 9.47 \pm19.14 $ & $ 23.05 \pm31.04 $ & $ \mathbf{4.43 \pm6.01} $& $ 7.3 \pm 17.53 $ 
						\\ \hline
						&     Reward     &   $  -11.46 \pm 1.35 $     & $ -11.94 \pm1.38 $ & $ -11.52 \pm1.1 $ & $ -16.92 \pm3.41 $ & $ \bm{-11.44 \pm 1.48} $ \\ 
						5	&    Average distance     &   $  1.26 \pm 0.27$     & $ 1.35 \pm0.29 $ & $ \mathbf{ 1.21 \pm0.22}$  &  $2.37 \pm0.68 $& $1.26 \pm 0.29 $  \\
						&	  \# collisions & $  13.88 \pm 19.94 $ & $ 16.94 \pm28.21     $ & $ 42.52 \pm36.86 $ & $ \mathbf{5.24 \pm6.53} $ & $ 13.73 \pm 22.98 $ 
						\\ \hline
						&     Reward     &   $ -18.23 \pm 2.48 $      & $ -19.07 \pm2.06 $ & $ -18.21 \pm2.24$ & $ -29.11 \pm5.46 $& $ \mathbf{-18.08 \pm 2.32} $ \\ 
						6	&    Average distance     &   $ 2.0 \pm 0.41  $     & $ 2.13 \pm0.35 $ & $ \mathbf{1.93 \pm0.37} $ & $ 3.83 \pm0.91  $ & $1.96 \pm 0.38 $  \\
						&	  \# collisions & $  23.43 \pm 26.02 $  & $ 30.72 \pm33.47 $ & $ 59.9 \pm30.03 $ & $ \mathbf{11.04 \pm10.11}  $& $ 31.06 \pm 35.49 $ 
						\\ \hline 
					\end{tabular}
				}
				\caption{{\color{black}Comparison of MADDPG, MA-MADDPG, CommNet, MAAC, and MD-MADDPG on Cooperative Navigation when increasing the number of agents.}}
				\label{tab:exp_cn_moreagents}
			\end{center}
		\end{table*}	
		
		\clearpage
		\subsection{Increasing the number of agents - Partial Observable Cooperative Navigation} \label{subsec:moreagents_POCN}
		Table \ref{tab:exp_pocn_moreagents} presents the comparison of {\color{black}MADDPG, MA-MADDPG, CommNet, MAAC and MD-MADDPG} on Partially Observable Cooperative Navigation when the number of agents increases. 
		It can be noted that MD-MADDPG still achieves good performance and in some scenarios (e.g. number of agents = 5) it outperforms other {\color{black} methods.}
		
		\begin{table*}[h]
			\begin{center}
				{\color{black}
					\begin{tabular}{lllllll}
						\textbf{\# of agents}	 &  \textbf{Metric}	&  \textbf{MADDPG} & \textbf{MA-MADDPG} & \textbf{CommNet} & \textbf{MAAC} & \textbf{MD-MADDPG}
						\\ \hline 
						&     Reward     &   $ \bm{-4.66 \pm 0.76} $      & $ -4.96 \pm 0.95 $ & $ -5.15 \pm 0.86$ & $ -6.73 \pm 1.44 $ & $ -4.97 \pm 0.94 $ \\ 
						3	&    Average distance     &   $ \bm{0.54  \pm 0.25}   $     & $  0.65 \pm 0.31 $ & $  0.7 \pm 0.29$ & $ 1.21 \pm 0.47 $ & $ 0.65 \pm 0.31 $  \\
						&	  \# collisions & $ 2.35 \pm 4.76 $  & $  1.54 \pm 4.5 $ & $   4.18 \pm 7.76$ & $ 9.61 \pm 13.28 $ & $ \bm{2.35 \pm 4.61} $ 
						\\ \hline 
						&     Reward     &   $ -8.11 \pm 1.37 $      & $ \mathbf{-7.56 \pm 1.17 } $ & $ -9.05 \pm 1.49 $ & $  -10.79 \pm 2.07 $ & $ -8.17 \pm 1.44 $ \\ 
						4	&    Average distance     &   $ 1.02 \pm 0.34   $     & $ \mathbf{0.88 \pm 0.29} $ & $  1.19 \pm 0.36 $ & $  1.68 \pm 0.51$ & $ 1.04 \pm 0.36 $  \\
						&	  \# collisions & $ 4.97 \pm 7.56 $  & $  \mathbf{2.82 \pm 5.47} $ & $ 29.5 \pm 28.71 $ & $  7.03 \pm 7.18$ & $ 2.89 \pm 5.24 $ 
						\\ \hline
						&     Reward     &   $ -16.33 \pm 3.06 $     & $ -16.56 \pm 2.53 $ & $ -15.79 \pm 2.61 $ & $  -16.59 \pm 3.11 $  & $ \bm{-15.29 \pm 3.07} $ \\ 
						5	&    Average distance     &   $ 0.66 \pm 0.18   $    & $ 1.68 \pm 0.5 $ & $ 1.48 \pm 0.51 $ & $  6.13 \pm 7.44$ & $\bm{0.61 \pm 0.18} $  \\
						&	  \# collisions & $  7.86 \pm 8.38 $  & $ 15.64 \pm 17.38 $ & $  67.01 \pm 44.81 $ & $ \mathbf{2.31 \pm 0.62}$ & $ 3.38 \pm 4.2 $ 
						\\ \hline
						&     Reward     &   $ -18.69 \pm 3.18 $      & $ -20.24 \pm 2.62 $ & $ \mathbf{-17.11 \pm 2.86}$ & $  -21.3 \pm 4.61 $ & $ -39.83 \pm 2.29 $ \\ 
						6	&    Average distance     &   $ 2.09 \pm 0.53   $    & $ 2.31 \pm 0.43  $ & $ \mathbf{1.72 \pm 0.47}$ & $  2.53 \pm 0.76$  & $5.63 \pm 0.38  $  \\
						&	  \# collisions & $  13.14 \pm 13.21 $  & $ 35.38 \pm 19.84 $ & $ 76.64 \pm 48.27 $ & $  8.99 \pm 8.63$  & $ \bm{6.47 \pm 6.1} $ 
						\\ \hline 
					\end{tabular}
				}
				\caption{{\color{black} Comparison of MADDPG, MA-MADDPG, CommNet, MAAC and MD-MADDPG on Partial Observable Cooperative Navigation when increasing the number of agents.}}
				\label{tab:exp_pocn_moreagents}
			\end{center}
		\end{table*}
		\vspace*{\fill}
		
		\clearpage
		\subsection{Ablation study} \label{subsec:ablatation_context}
		In the experiments presented here we study the benefits of each specific component, such as the context vector and the modules required for communicating, on final performance.
		To assess the importance of the context vector (Eq. \ref{eq:readh}) we have run a set of experiments removing $ \mathbf{h}_i $ from the reading module of the agents. Table \ref{tab:nocontext} shows that by using only $ \mathbf{e}_i $ without $ \mathbf{h}_i $ during the reading phase, the performance overall degrades on {\color{black} almost all the environments. We have noticed that the role played by the context vector becomes more critical as the level of communication requires by the underlying task increases, like in Sequential CN, Synchronous CN and Waterworld.
			We also run experiments to assess the performance of the components involved in the functioning of communication. It resulted that removing either the reading or the writing modules the performance significantly worsened on every scenario.
		}

		\begin{table*}[h]
			\begin{center}
				{\color{black}
					\begin{tabular}{llllll}
						\textbf{Environment}		&  \textbf{Metric} & \textbf{no context}& \textbf{no read} & \textbf{no write} &\textbf{MD-MADDPG} 
						\\ \hline 
						&     Reward     &      $ -2.28 \pm 0.1 $ & $-35.13 \pm2.07$ & $ -9.04 \pm5.37 $  & $ \bm{-2.27 \pm 0.10}  $\\ 
						CN	&    Average distance     &      $  0.14 \pm 0.05 $& $16.57 \pm1.04$ & $ 3.51 \pm2.68 $  &$ \bm{0.13  \pm 0.05}$\\
						&	  \# collisions  & $ \bm{0.08 \pm 0.59} $& $0.2 \pm0.96$ & $ 1.52 \pm3.32 $  &$ 0.12     \pm 0.82$
						\\ \hline 
						&     Reward     &      $ -2.68 \pm 0.45 $ & $-5.4 \pm0.75$  & $ -5.93 \pm0.07 $  & $ -2.68  \pm 0.46  $ \\ 
						PO-CN	&    Average distance     &      $ 0.34 \pm 0.22  $& $ 1.7 \pm0.38 $   & $ 1.96 \pm0.03 $  & $0.34 \pm 0.22 $ \\
						&	  \# collisions  & $ 0.32 \pm 1.14 $& $0.56 \pm1.61$ & $ 0.58 \pm1.56 $  & $ \bm{0.26  \pm 1.06} $
						\\ \hline
						&     Reward       &      $ \bm{189.8 \pm 38.39}  $& $ -16.5 \pm2.69 $  &  $ -16 \pm0.1 $ &  $ 92.90 \pm 69.78  $\\ 
						Sync CN	&  \# sync occup.        &     $ \bm{51.61 \pm 58.65} $& $ 0.1 \pm0.72 $  & $ 0.2 \pm0.92 $  &   $ 31.6 \pm 19.34 $\\ 
						&	  \# not sync occup.  & $ 103.36 \pm 61.23 $& $ 21.22 \pm59.75 $  & $ 2.68 \pm2.9 $  & $ \bm{17.58 \pm 12.00}  $
						\\ \hline
						Sequential CN 
						&     Reward       &   $113.33 \pm 48.11			  $ & $ -13.59 \pm0.77 $  & $ -13.85 \pm0.12 $  &  $ \bm{130.15  \pm 35.19}  $  \\ 
						&     Average distance        &   $  129.79 \pm 83.56 $& $ 377.73 \pm35.6$  & $ 391.75 \pm5.87 $  &  $ \bm{99.15 \pm 50.59} $
						\\ \hline
						Swapping CN &     Reward      &      $ 75.76 \pm 66.49 $& $ -15.53 \pm0.37 $  &$ -12.54 \pm2.52 $   & $\bm{129.63 \pm 47.26 } $\\ 
						&     Average distance        &   $ 158.39 \pm 108.42 $& $ 579.23 \pm10.02 $  & $ 416.8 \pm86.15 $  & $ \bm{53.21  \pm 40.80} $
						\\ \hline
						&     Reward     &    $ 31.31 \pm 77.31  $ & $ -1.5 \pm2.33 $  & $ 21.95 $  & $\bm{503.96} \pm 103.91  $ \\ 
						Waterworld	&    \#  food targets            &   $ 1.8 \pm 4.04 $ & $ 0.21 \pm 0.11 $  & $  1.18 \pm1.12$  & $ \bm{26.25 \pm 5.41}  $ \\
						&	  \# poison targets  & $ \bm{5.22 \pm 2.95} $& $ 3.4 \pm2.22 $  & $  3.98 \pm2.28 $  & $ 7.77 \pm 2.95 $
					\end{tabular}
					\caption{An assessment of MD-MADDPG without context vector, MD-MADDPG without reading operation, MD-MADDPG without writing operation compared with the standard version MD-MADDPG.}
					\label{tab:nocontext}
				}
			\end{center}
		\end{table*}	
		\vspace*{\fill}

		\clearpage
		\subsection{Main results visualization} \label{subsec:mainresultsvisualization}
		{\color{black} 
			Figure \ref{fig:all_results_visualization} provides boxplot visualization of the main results already presented in Table \ref{tab:all_results}.
			
			\begin{figure}[h]
				\subfloat[Cooperative Navigation]{\includegraphics[width = 0.50\linewidth]{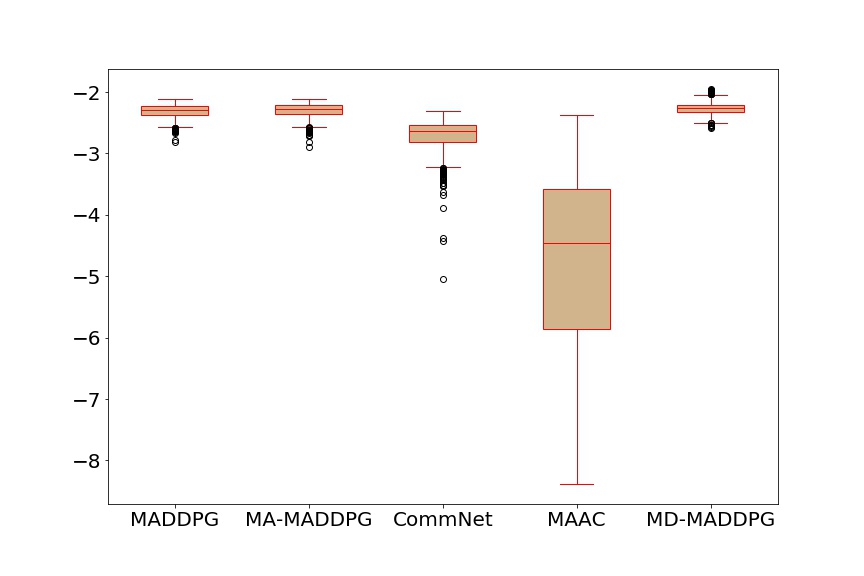}} 
				\subfloat[Partial Observable CN]{\includegraphics[width = 0.50\linewidth]{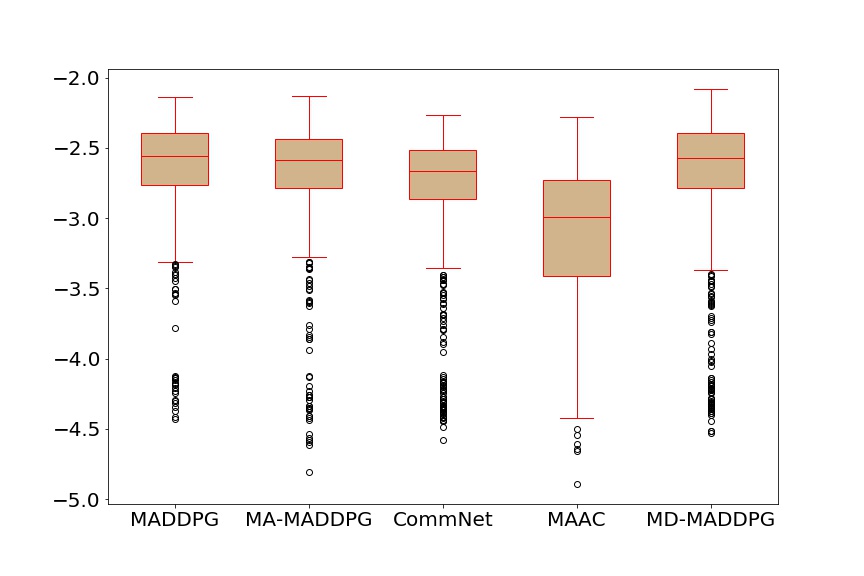}}
				\\
				\subfloat[Synchronous CN]{\includegraphics[width = 0.50\linewidth]{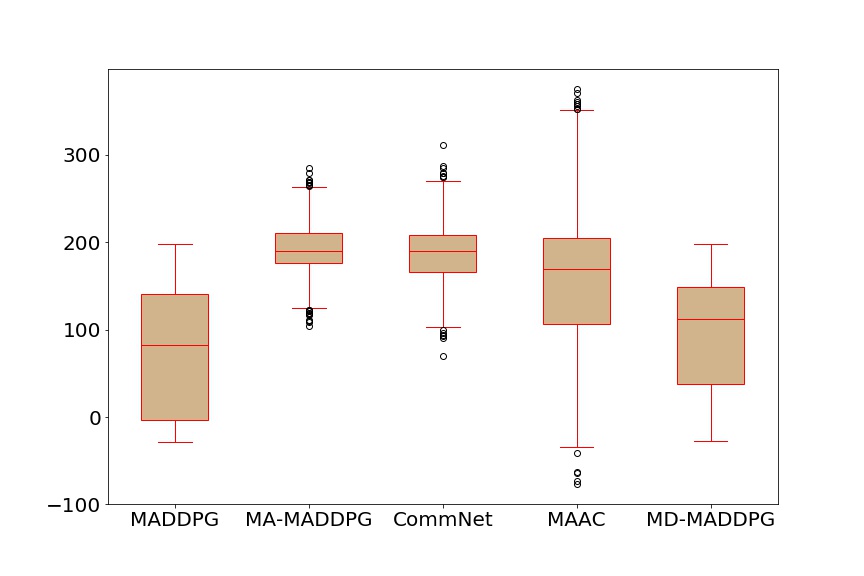}}
				\subfloat[Sequential CN]{\includegraphics[width = 0.50\linewidth]{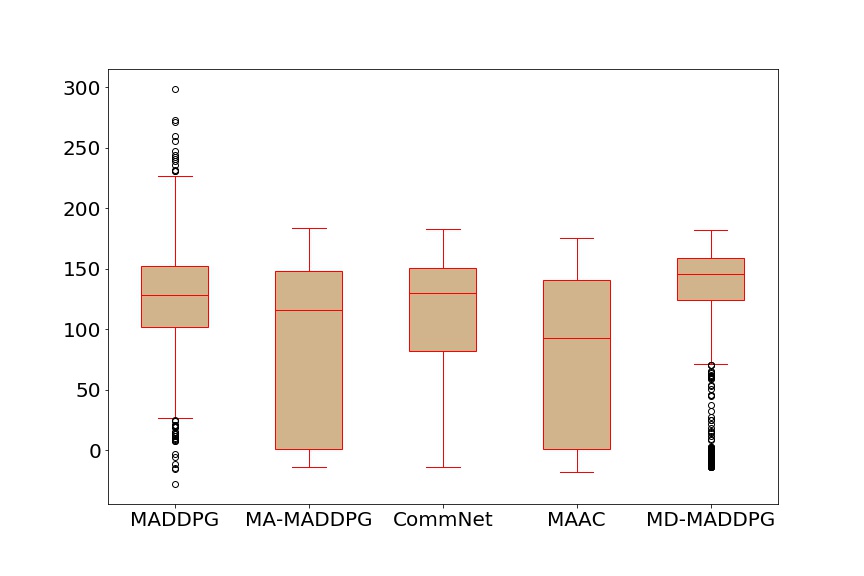}} 
				\\
				\subfloat[Swapping CN]{\includegraphics[width = 0.50\linewidth]{img_res_simple_spread_swap.jpg}}
				\subfloat[Waterworld]{\includegraphics[width = 0.50\linewidth]{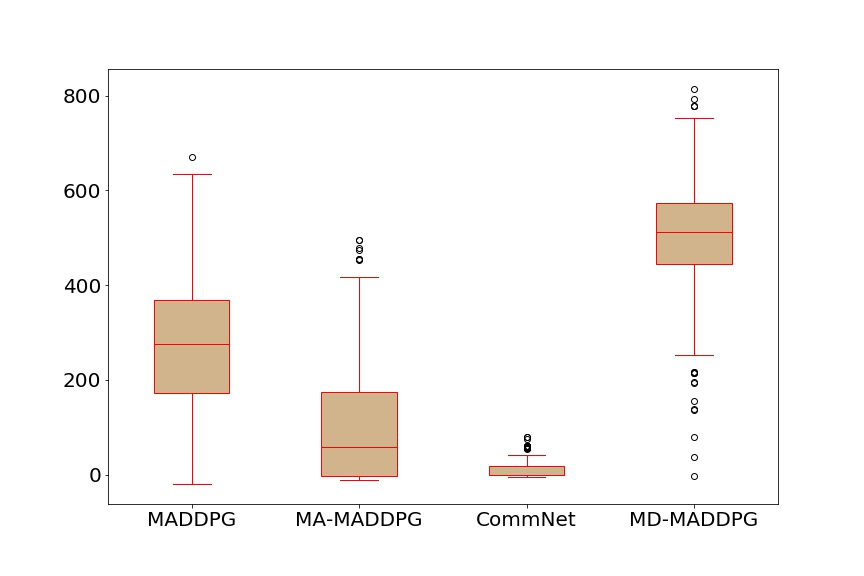}}
				\caption{{\color{black}Boxplot representing the main results in Table \ref{tab:all_results}.}}
				\label{fig:all_results_visualization}
			\end{figure}
			
			\clearpage	
			\subsection{Multiple seeds} \label{subsec:multipleseeds}
			In this section, we investigate the sensitivity of MD-MADDPG on changes in random seeds used for setting the initial conditions of the randomness in the learning process. To show that the presented results are not affected by a particular choice of the seed that can significantly condition the final performance, we report the outcome of varying different seeds. 
			Figures \ref{fig:simple_spread_swap_multiple_seeds} and \ref{fig:simple_spread_sequential_multiple_seeds} show respectively the results of MD-MADDPG on Swapping CN and Sequential CN when changing the seed for training and testing the model. It can be noted that in both cases, models are not seed-sensitives, indeed varying the seed does not significantly affect the final results.
			In order to investigate the statistical significance of these results, we carried out a MANOVA (Multivariate ANOVA) \citep{french2008multivariate}, assessing the null hypothesis that all the population means are the same. In both scenarios, there is no enough evidence to conclude that there is a difference in means at the $0.001 $ significance level ($ p $-values is $ 0.1267 $ on Swapping CN and $ 0,8357 $ on Sequential CN).
			
			\begin{figure}[h]
				\centering
				\includegraphics[scale=0.4]{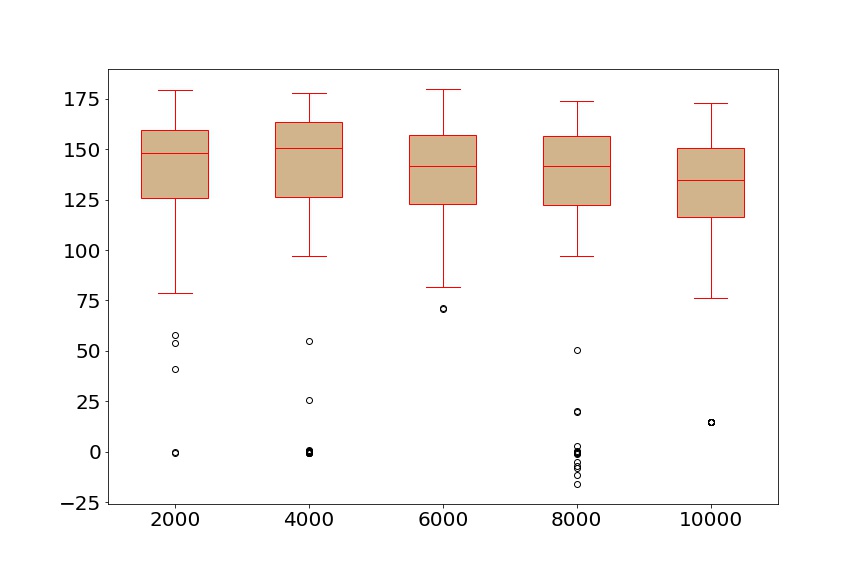}
				\caption{{\color{black}Boxplot representing the results of MD-MADDPG on Swapping CN when changing different seeds. The horizontal axis shows the seed and the vertical axis the reward. The MANOVA test returned a $ p $-value of $ 0.1267 $. This confirms that there is no enough evidence to conclude that the means are different.} }
				\label{fig:simple_spread_swap_multiple_seeds}
			\end{figure}		
			
			\begin{figure}[h]
				\centering
				\includegraphics[scale=0.4]{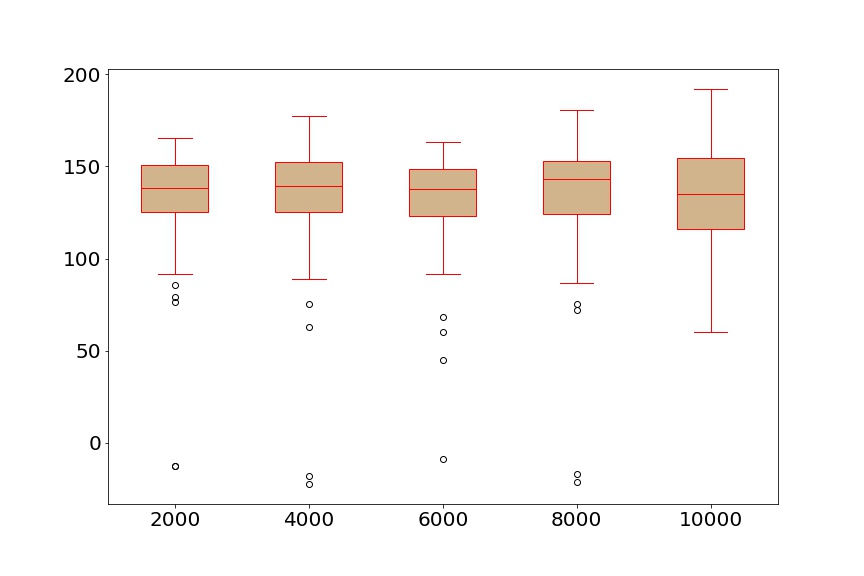}
				\caption{{\color{black}Boxplot representing the results of MD-MADDPG on Sequential CN when changing different seeds. The horizontal axis shows the seed and the vertical axis the reward. The MANOVA test returned a $ p $-value of $ 0.8357 $. This confirms that there is no enough evidence to conclude that the means are different.}}
				\label{fig:simple_spread_sequential_multiple_seeds}
			\end{figure}	
			
			\clearpage
			\subsection{Multiple memory sizes} \label{subsec:memorysizes}
			Figure \ref{fig:simple_spread_swap_memory_sizes} represents how the memory size affects the resulting reward on Swapping Cooperative Navigation.
			It can be noted that given the selected architecture, the higher reward is obtained using a memory size of 200.
			
			\begin{figure}[h]
				\centering
				\includegraphics[scale=0.4]{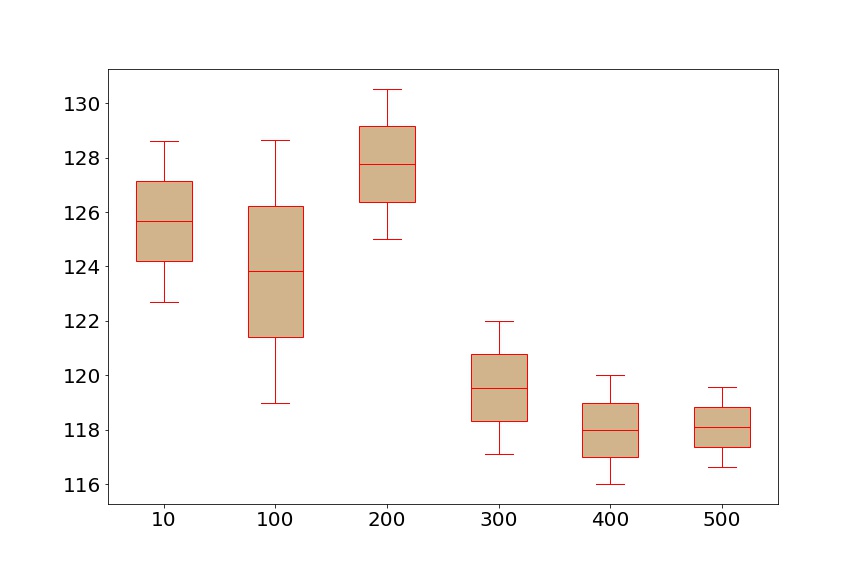}
				\caption{{\color{black}Results obtained on Swapping CN using different memory dimensions. The horizontal axis report the memory size and the vertical axis the reward.}}
				\label{fig:simple_spread_swap_memory_sizes}
			\end{figure}

			\clearpage
			\subsection{Communication analysis - Read Vector}\label{subsec:readvector}
			Figure \ref{fig:memory_PC3_read} shows the results of a communication analysis analogous to Section \ref{sec:commanalysis} but for the read vector.
			As for the write vector, communication patterns emerge and seem to point out that communication is more intense when coordination is highly required. 
			It can be noted that the read vectors still correlate with the phases. For example, in Synchronous Cooperative Navigation, the first principal component is highly activated during phases 1 and 2. This suggests that agents intensely communicate to reach the landmarks simultaneously. 
			A different behaviour emerges for others environments like Swapping Cooperative Navigation where the reading vector is highly activated during the first phase, probably because the agents received the information about the next landmark to swap (e.g. coordinates). This analysis has been conducted to better present the behaviour of the agents and the interactions with their internal components. We believe that the content of the message, which corresponds to the write vector, is more informative since it is what the agents are explicitly communicating. On the other side the read vector can be more difficult to interpret since it is internally used by the agents together with other components to achieve communication. 
			\begin{figure*}[h] 
				\begin{minipage}[b]{0.45\linewidth}
					\centering
					\includegraphics[width=1\linewidth]{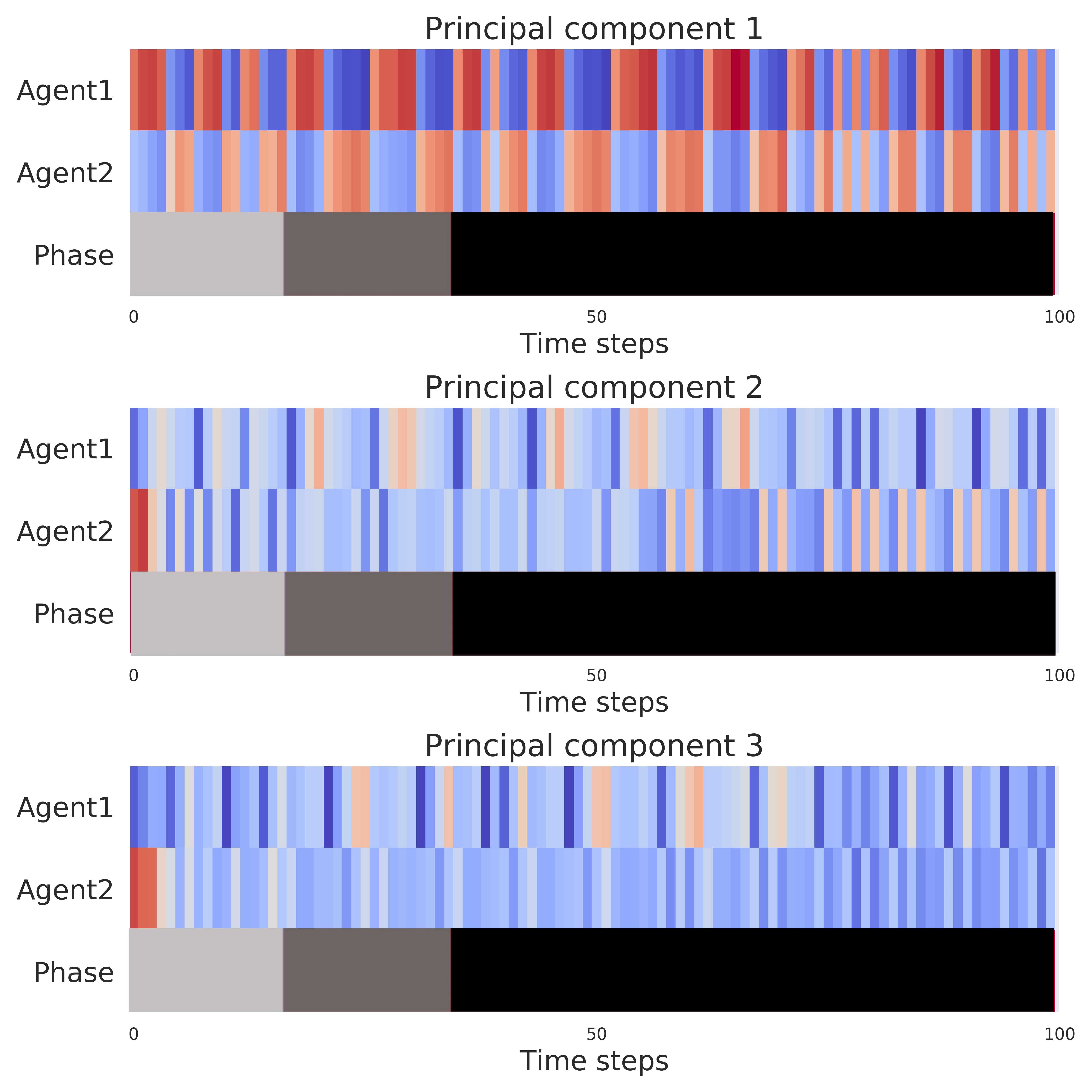} 
					\\ (a) Sequential CN
					\vspace{1ex}
				\end{minipage}
				\begin{minipage}[b]{0.45 \linewidth}
					\centering
					\includegraphics[width=1\linewidth]{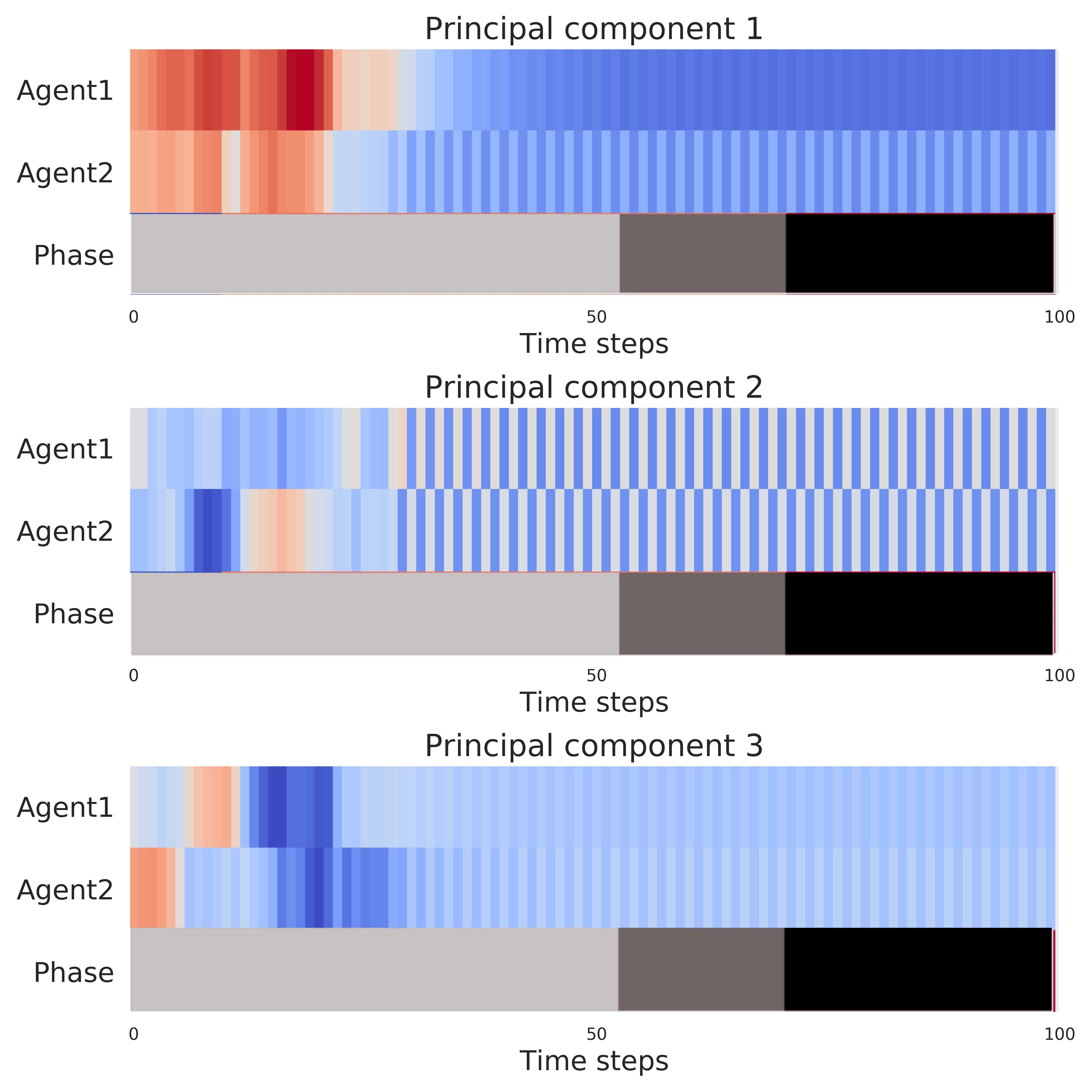} 
					\\ (b) Swapping CN
					\vspace{1ex}
				\end{minipage} 
				\\
				\begin{minipage}[b]{0.45\linewidth}
					\centering
					\includegraphics[width=1\linewidth]{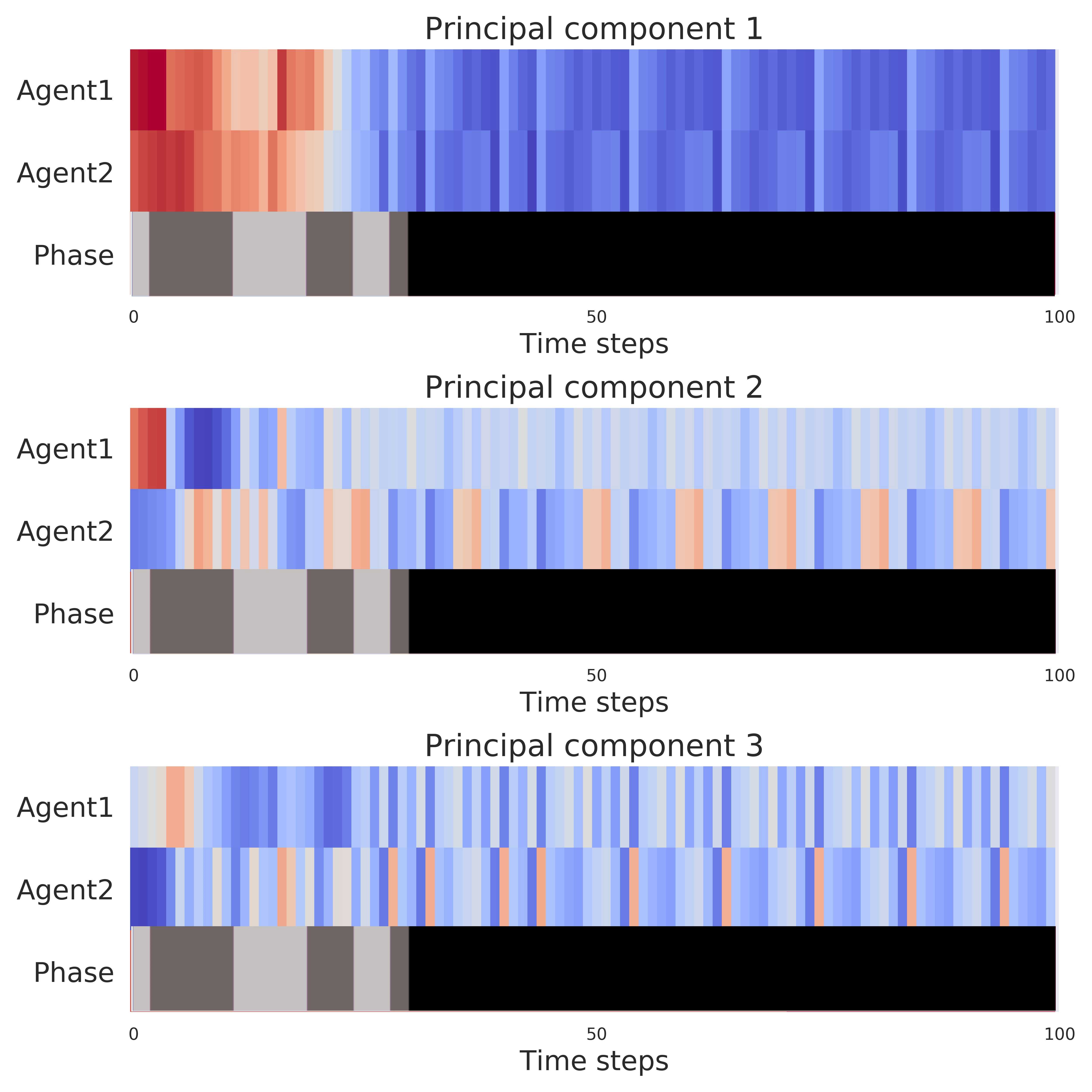} 
					\\ (c) Synchronous CN
					\vspace{1ex}
				\end{minipage} 
				\begin{minipage}[b]{0.45\linewidth}
					\centering
					\includegraphics[width=1\linewidth]{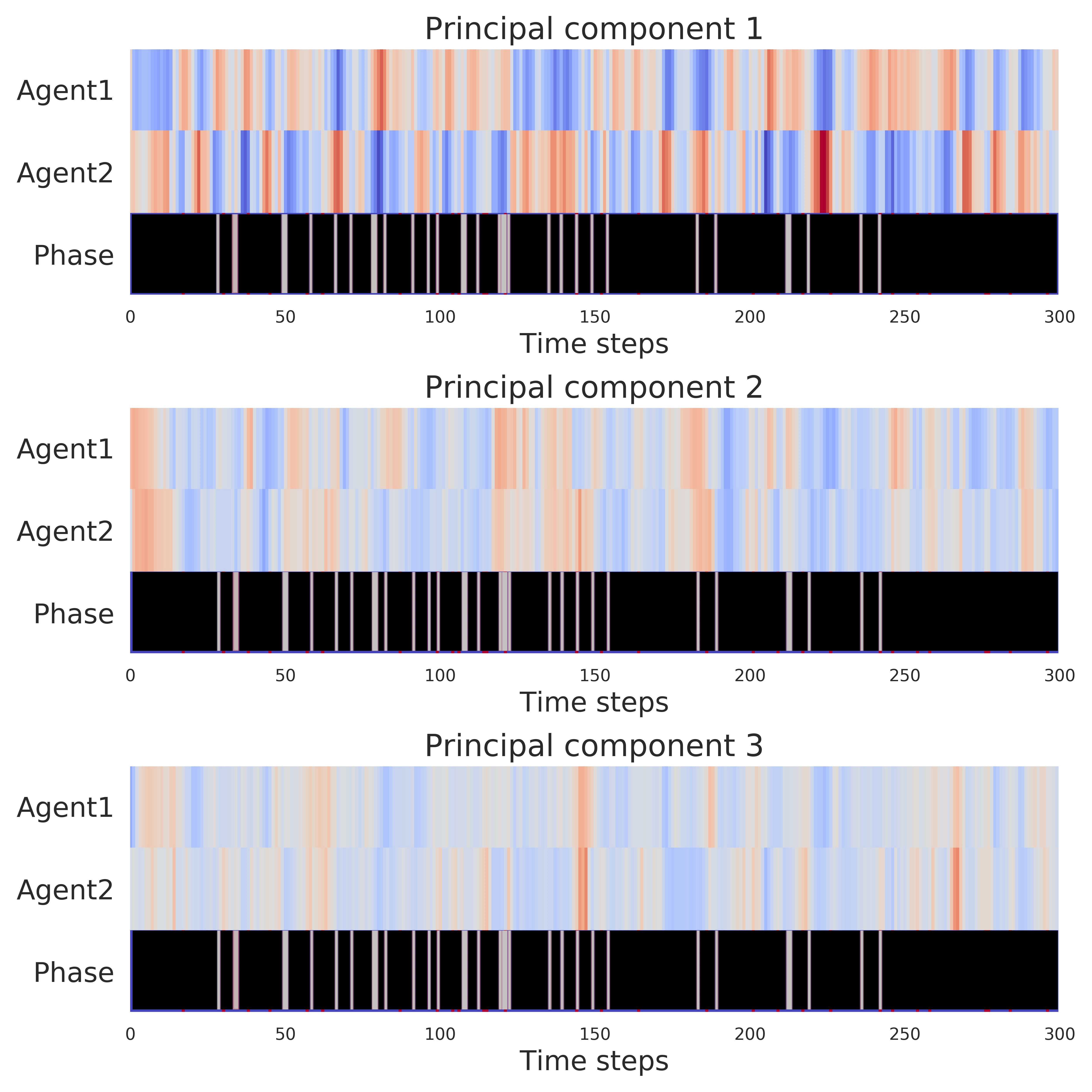} 
					\\ (d) Waterworld
					\vspace{1ex}
				\end{minipage}
				
				\caption{{\color{black}Visualisation of communication strategies learned by the agents in four different environments: the three principal components provide orthogonal descriptors of the read vector content of the agents and are being plotted as a function of time. Within each component, the highest values are in red, and the lowest values are in blue. The bar at the bottom of each figure indicates which phase (or sub-task) was being executed within an episode.}
				}
				\label{fig:memory_PC3_read} 
			\end{figure*}
			
		}
	\end{document}